\documentclass[10pt,authoryear]{article}%

\usepackage{qcircuit}  
\RequirePackage{amsthm,amsmath,amsfonts,amssymb}
\RequirePackage[numbers]{natbib}
\usepackage{graphicx} 
\RequirePackage[colorlinks,citecolor=blue,urlcolor=blue]{hyperref}
\usepackage{algorithm}
\usepackage{algorithmic}

\theoremstyle{italiclabel}
\newtheorem{theorem}{Theorem}
\newtheorem{definition}[theorem]{Definition}
\newtheorem{proposition}[theorem]{Proposition}

\allowdisplaybreaks

\begin{document}
	\title{Robust Iterative Learning Hidden Quantum Markov Models}
	\author{Ning Ning\thanks{Department of Statistics,
			Texas A\&M University, College Station, Texas, USA. Email: patning@tamu.edu.}}
	\date{}
	\maketitle

\begin{abstract}
Hidden Quantum Markov Models (HQMMs) extend classical Hidden Markov Models to the quantum domain, offering a powerful probabilistic framework for modeling sequential data with quantum coherence. However, existing HQMM learning algorithms are highly sensitive to data corruption and lack mechanisms to ensure robustness under adversarial perturbations. In this work, we introduce the Adversarially Corrupted HQMM (AC-HQMM), which formalizes robustness analysis by allowing a controlled fraction of observation sequences to be adversarially corrupted. To learn AC-HQMMs, we propose the Robust Iterative Learning Algorithm (RILA), a derivative-free method that integrates a Remove Corrupted Rows by Entropy Filtering (RCR-EF) module  with an iterative stochastic resampling procedure for physically valid Kraus operator updates. RILA incorporates L1-penalized likelihood objectives to enhance stability, resist overfitting, and remain effective under non-differentiable conditions. Across multiple HQMM and HMM benchmarks, RILA demonstrates superior convergence stability, corruption resilience, and preservation of physical validity compared to existing algorithms, establishing a principled and efficient approach for robust quantum sequential learning. 

\end{abstract}

%
%


\section{Introduction}

We first provide the background in Section~\ref{sec:background}, present the motivation in Section~\ref{sec:motivation}, and summarize our main contributions in Section~\ref{sec:Contributions}, followed by the overall organization of the paper in Section~\ref{sec:Organization}.

\subsection{Background}
\label{sec:background}
Classical Hidden Markov Models (HMMs) are probabilistic graphical models widely used to analyze sequential data in dynamic systems, where state transitions follow the Markov property, meaning each future state depends solely on the current state, independent of prior states. These models represent an unobserved latent process that evolves over time, generating observable outputs probabilistically at each step, with belief states which are probability distributions over latent states, providing an interpretable depiction of system dynamics. HMMs have become foundational in temporal modeling, with applications in fields like speech recognition and bioinformatics. Extending this framework, Hidden Quantum Markov Models (HQMMs), introduced by \citet{monras2010hidden}, incorporate quantum mechanics by replacing classical probability-vector belief states with quantum density matrices and using symbol-conditioned completely positive maps (Kraus operators) to govern state transitions and emissions. 

The quantum parameterization enables HQMMs to capture complex dependencies and correlations that surpass classical probabilistic structures, often requiring fewer internal quantum states to model sequence dependencies compared to classical HMMs. As a result, HQMMs offer a more expressive framework for modeling sequential processes influenced by quantum effects or intricate latent dynamics. Early low-dimensional HQMMs \citep{oneill2012toy} demonstrated compact examples contrasting classical and quantum correlations. \cite{clark2015open} recast HQMMs as open quantum systems with instantaneous feedback, linking the framework to physically realizable dynamics. 	
\cite{cholewa2017quantum} extended the idea of quantum Markov chains to propose quantum hidden Markov models (QHMMs). 
Further analyses explored expressiveness: \citet{adhikary2020expressiveness} related HQMMs to observable operator models (OOMs) and showed that HQMMs form a positive subclass of OOMs, thus avoiding negative probabilities. Variants such as the circular HQMM \citep{javidian2021chqmm} and the split HQMM \citep{li2023shqmm} expand representation power using tensor-network or conditional master-equation structures. 

Learning HQMMs is considerably more challenging than training classical HMMs due to the quantum nature of their latent state representation. The hidden state in an HQMM is a density matrix that evolves through completely positive trace-preserving (CPTP) maps parameterized by Kraus operators, which must satisfy nonlinear trace-preserving constraints~\citep{monras2010hidden}. This introduces nonconvexity, non-commutativity, and severe identifiability issues during optimization~\citep{srinivasan2018hilbert, adhikary2020expressiveness}. Moreover, no closed-form Expectation–Maximization (EM) updates exist for HQMMs, making likelihood maximization computationally expensive~\citep{clark2015open}. To address these challenges, several computational strategies have been developed, including projected stochastic gradient descent for constrained learning~\citep{srinivasan2018learning}, variational or neural approximations to parameterize quantum channels~\citep{adhikary2020expressiveness},  and tensor-network formulations for efficient quantum state propagation~\citep{javidian2021chqmm}, specifying a QHMM learning algorithm as adaptive evolutionary search in the space of quantum circuits with mid-circuit measurement \citep{markov2022implementation}. 

\subsection{Motivation}
\label{sec:motivation}
Learning the properties of quantum states is a central challenge in quantum computing. Most studies assume that measurements are precise and align perfectly with their mathematical formulations. In practice, this assumption often fails, particularly in the current noisy intermediate-scale quantum (NISQ) era~\citep{preskill2018quantum}, where state preparation and measurement (SPAM) errors are common due to imperfect devices. To address this, prior works have explored robustness to SPAM errors~\citep{brandao2020fast, yu2023robust, stilck2024efficient, aliakbarpour2025adversarially}. However, SPAM noise may not capture all possible errors in a quantum system. Unforeseen physical factors, such as an unstable power supply or vibrations from a subway train passing beneath the laboratory, can introduce additional disruptions. Furthermore, human factors may contribute, ranging from benign errors, such as misreading measurements, to malicious acts, such as a knowledgeable adversary deliberately sabotaging the system. Consequently, there is a critical need to develop a more robust corruption model for quantum state learning that accounts for these unpredictable factors and to design algorithms that remain effective in this challenging setting. 

In \cite{aliakbarpour2025adversarially}, the authors introduce the adversarial corruption model, where an adversary can arbitrarily corrupt a $\gamma$-fraction of measurement outcomes. In the standard setup, there are $n$ copies of a quantum state $\rho$  and corresponding measurements $\mathcal{M}^n = (\mathcal{M}_1, \ldots, \mathcal{M}_n)$, where $\mathcal{M}_i = \{M_o^i\}$, are applied to each copy, with the outcome of measuring the $i$th copy using $\mathcal{M}_i$ denoted by $o_i$. The $\gamma$-adversarial corruption model is defined as follows: First, measurements are applied to all copies of $\rho$, yielding outcomes $(o_1, \ldots, o_n)$. Then, the adversary, with full knowledge of the measurements used, arbitrarily corrupts a $\gamma$-fraction of these outcomes. Finally, the algorithm uses the corrupted outcomes $(y_1, \ldots, y_n)$ to learn properties of the state $\rho$. The authors then demonstrated that $\gamma$-adversarial corruption is a stronger model than SPAM noise. Under this more robust corruption model, they developed an algorithm employing non-adaptive measurements that can learn an unknown rank-$r$ quantum state to within $\tilde{O}(\gamma\sqrt{r})$ in trace distance, provided the number of state copies is sufficiently large.

The HQMM operates in a much more complex setting. On one hand, it serves as a quantum graphical model that encodes dependencies among quantum systems in a graphical structure; on the other hand, HQMMs are designed to model sequential data, where each observation depends (directly or indirectly) on past observations through some hidden internal state. This dual nature complicates the standard corruption framework. Nevertheless, as a quantum model, an HQMM remains susceptible to various sources of error, such as environmental noise, hardware imperfections, and deliberate adversarial perturbations that exploit quantum vulnerabilities. Developing robust HQMMs under adversarial corruption is therefore essential for ensuring reliable performance in real-world quantum computing applications, where unexpected errors or malicious manipulations could otherwise compromise state estimation and prediction accuracy. Once such a notion of adversarially robust HQMMs is properly formulated, several natural research questions emerge:
\begin{itemize}
	\item How can we design a corruption-robust mechanism that is broadly adaptable to HQMM learning algorithms and maintains reliability under adverse conditions? 
	\item Can we construct quantum learning algorithms that remain robust under arbitrary corruptions while ensuring compatibility with multiple evaluation metrics across diverse scenarios? 
	\item Beyond robustness, can we fundamentally improve existing learning procedures so that they consistently yield valid quantum operators without approximation errors?
\end{itemize}
Addressing these questions constitutes the central goal of this paper.

\subsection{Our Contributions}
\label{sec:Contributions}
We introduce the concept of \emph{$\gamma$-adversarial corruption} to the HQMM framework by defining the Adversarially Corrupted HQMM (AC-HQMM) in Section \ref{sec:AC_HQMM}. This formulation extends the conventional HQMM by allowing a fraction $\gamma$ of the generated observation sequences to be arbitrarily corrupted by an adversary with full knowledge of the measurement process. The AC-HQMM thus provides a principled mechanism to model and analyze the robustness of quantum sequential learning systems under adversarial perturbations, bridging the gap between quantum probabilistic modeling and robust learning theory. Notably, when $\gamma = 0$, the proposed model reduces exactly to the standard HQMM, preserving its probabilistic consistency while enabling controlled analysis of corruption effects.

To address the challenge of learning HQMMs under adversarial corruption, we propose the Robust Iterative Learning Algorithm (RILA) for AC-HQMMs. RILA builds upon the Iterative Learning Algorithm (ILA) proposed in \cite{srinivasan2018learning}, which remains the only known method guaranteeing valid quantum operators while avoiding restrictive rank assumptions and hyperparameter tuning. 
 These advantages make ILA a natural foundation for robustness-oriented extension. However, existing likelihood-based HQMM learning frameworks are highly sensitive to corrupted or anomalous data, which can substantially bias parameter estimation and degrade model performance. To mitigate this issue, we introduce a corruption-aware preprocessing module, termed RCR-EF (Remove Corrupted Rows by Entropy Filtering), that identifies and removes corrupted sequences from the training matrix based on multiple statistical indicators, including Shannon entropy, unique value count, absolute mean deviation, and variance. This step is specifically designed for spatiotemporal quantum data, where cross-sequence dependencies invalidate classical i.i.d.-based filtering techniques.

The complete RILA framework integrates this robust filtering step with a batched, iterative likelihood-maximization process and a sequential resampling mechanism for parameter refinement. Within each batch, RILA generates multiple candidate updates to the Kraus operator matrix and evaluates their physical validity and log-likelihood. A stochastic resampling step, weighted by exponentiated log-likelihoods, then probabilistically selects the most promising candidate, ensuring both exploitation of high-performing updates and exploration of the broader parameter space. This design enhances the performance of local optimizers by introducing controlled stochasticity, enabling the algorithm to escape local minima without resorting to computationally expensive global optimization methods. Importantly, RILA achieves these robustness and efficiency improvements without increasing the number of Kraus operators, thereby maintaining the computational scalability of ILA while significantly improving stability and accuracy under corruption and noise.

Ensuring robustness under corruptions while maintaining compatibility with multiple evaluation metrics is crucial for reliable HQMM learning in practical quantum systems. Absolute value (L1) penalties are widely used in statistical and machine learning models to encourage sparsity, stabilize parameter estimates, and reduce sensitivity to outliers or model misspecification. In the HQMM setting, Kraus operator parameters can grow large as model complexity increases, and incorporating L1-type penalties into the log-likelihood objective mitigates overfitting while enhancing robustness to corrupted or noisy observations. Such penalties introduce non-differentiable components into the objective, rendering gradient-based methods less effective; derivative-free optimizers like the pattern search algorithm are particularly well-suited for handling these challenges. RILA leverages this principle by combining L1-penalized log-likelihoods with a sequential resampling mechanism, which probabilistically selects among multiple candidate updates based on their weighted log-likelihoods. This approach not only improves exploration of the parameter space and convergence efficiency but also ensures that the learned HQMM parameters remain physically valid, yielding stable and reliable performance across diverse scenarios and evaluation metrics. 

To comprehensively assess performance, we evaluate RILA under two likelihood objectives: the standard (regular) log-likelihood and a penalized (irregular) variant that introduces regularization terms to discourage overfitting and ensure physically valid operators. This results in a $2 \times 2$ experimental grid:
\begin{itemize}
	\item regular likelihood on clean data,
	\item penalized likelihood on clean data,
	\item regular likelihood on corrupted data,
	\item penalized likelihood on corrupted data.
\end{itemize}
The evaluation is conducted across three benchmark tasks: a 2-state 4-output HQMM from~\cite{monras2010hidden}, a 2-state 6-output HQMM from~\cite{srinivasan2018learning} , and a classical 8-state 8-output HMM. Each benchmark task is tested under all four conditions of the experimental grid.
Through this systematic comparison, we demonstrate RILA’s superior convergence stability, robustness to corruption, and preservation of physical validity. RILA consistently outperforms ILA across all tested configurations, often achieving high log-likelihoods with few training batches, whereas increasing the number of batches for ILA could provide minimal or no improvement. Furthermore, both RILA and ILA significantly outperform the EM algorithm in all experiments involving data generated from fully quantum HQMMs. For datasets generated by classical HMMs, the classical EM algorithm is preferred over the RILA, as RILA's performance is hindered by learning unnecessary quantum coherence and enforcing zero off-diagonal entries in the state density matrix increases algorithmic complexity. However, when datasets are corrupted, RILA's robust learning capability allows it to achieve comparable performance to the EM algorithm.

\subsection{Organization of the paper}
\label{sec:Organization}
The rest of the paper proceeds as follows. Section~\ref{HQMM} introduces HQMMs, building upon the foundations of classical HMMs and the formalism of quantum states and operations. It explores their theoretical underpinnings and extends the HQMM framework to model potential adversarial perturbations. Section~\ref{LHL} focuses on likelihood-based learning for HQMMs, proposing RILA and assessing its compatibility with multiple evaluation metrics while ensuring quantum operator validity. Section~\ref{NA} presents a comprehensive numerical analysis, comparing RILA with the baseline ILA and the classical EM method across three benchmark HQMM learning tasks. Section~\ref{Conclusion} summarizes the key findings and outlines broader implications and future research directions. The proof of Proposition \ref{thm:diagonal_match} is deferred to Appendix \ref{proof:thm:diagonal_match}.

\section{Hidden Quantum Markov Models}
\label{HQMM}
This section builds on the foundational concepts of HMMs, presented in Section~\ref{sec:HMM}, and quantum states and operations, introduced in Section~\ref{sec:QSQO}. We investigate the theoretical foundations of HQMMs in Section~\ref{sec:HQMM}, their generalization of classical HMMs in Section~\ref{sec:HQMM_HMM}, and introduce their new modeling under adversarial conditions in Section~\ref{sec:AC_HQMM}, offering a comprehensive analysis of their capabilities for modeling complex sequential data.


\subsection{Hidden Markov Models}
\label{sec:HMM}
HMMs are probabilistic graphical models that describe dynamic systems with Markovian state transitions, where each future state depends only on the current one and not on the past. The belief states in HMMs, probability distributions over the latent states, offer a natural and interpretable representation of system dynamics, establishing HMMs as foundational tools for temporal modeling.
An $n$-dimensional HMM with a discrete observation set of size $m$ is defined as the tuple $(\mathbb{R}_{\geq 0}^n, \mathbf{A}, \mathbf{C}, \vec{x}_0)$. The non-negative initial belief state $\vec{x}_0$ represents a probability distribution over $n$ hidden states, normalized such that $\mathbf{1}^T \vec{x}_0 = 1$ with $\mathbf{1}$ being the all-ones column vector. The non-negative column-stochastic transition matrix $\mathbf{A} \in \mathbb{R}^{n \times n}_{\geq 0}$, satisfying $ \mathbf{1} ^T\mathbf{A}= \mathbf{1}^T$, governs the evolution of the hidden state via 
$$\vec{x}_t = \mathbf{A} \vec{x}_{t-1}.$$ The  non-negative column-stochastic emission matrix $\mathbf{C} \in \mathbb{R}^{m \times n}_{\geq 0}$ with $ \mathbf{1} ^T\mathbf{C}= \mathbf{1}^T$, models the likelihood of observations through $$\vec{y}_t = \mathbf{C} \vec{x}_t.$$

Upon observing data point $y_t$ at time $t$, Bayesian updating yields the posterior belief 
$$\vec{x}_t' = \frac{\operatorname{diag}(\mathbf{C}{(y_t,:)}) \vec{x}_t}{1^T \operatorname{diag}(\mathbf{C}{(\vec{y}_t,:)}) \vec{x}_t},$$ 
where $\text{diag}(\mathbf{C}_{(y_t,:)})$ is a diagonal matrix with the entries of the $y_t$-th row of $\mathbf{C}$ along the diagonal, and the denominator renormalizes the vector $\vec{x}_t'$ to a valid probability distribution. This process enables not only inference over hidden states but also computation of sequence likelihoods 
$$P(\bar{y}) = 1^T \operatorname{diag}(\mathbf{C}{(y_t,:)}) \mathbf{A} \cdots \operatorname{diag}(\mathbf{C}{(y_1,:)}) \mathbf{A} \vec{x}_0$$
 for an observation sequence $\bar{y} = y_1, \dots, y_t$. 
Instead of using the matrices $\mathbf{A}$ and $\mathbf{C}$ separately, we can define an observation-specific transition operator 
\begin{align}
	\label{eqn:T}
	\mathbf{T}_y = \operatorname{diag}(\mathbf{C}{(y,:)}) \mathbf{A}
\end{align}
 for each $y$, which compactly expresess the belief update  and the likelihood of the observation sequence respectively as
$$
\vec{x}_t = \frac{\mathbf{T}_{y_t} \vec{x}_{t-1}}{{1}^T \mathbf{T}_{y_t} \vec{x}_{t-1}}\quad\text{and}\quad P(\bar{y}) = {1}^T \mathbf{T}_{y_t} \cdots \mathbf{T}_{y_1} \vec{x}_0.
$$
Further relaxing the non-negativity constraints on the entries of $\vec{x}_0$, $\mathbf{A}$, and $\mathbf{C}$, while retaining the normalization and stochasticity conditions where applicable, yields the Observable Operator Model (OOM) proposed in \cite{jaeger2000observable}, a more general linear dynamical system that can model signed or negative probabilities in belief states.

\subsection{Quantum States and Quantum Operations}
\label{sec:QSQO}
Unlike the classical latent state $\vec{x}_t$ in an HMM, which represents a probability distribution over system states with non-negative entries summing to 1, a quantum state $|\psi\rangle$ is a column vector in a complex Hilbert space, where each entry is a probability amplitude corresponding to a system state. The probability of observing a particular state is given by the squared norm of its amplitude, and the sum of the squared norms across all states must equal 1 to conserve probability. For instance, the quantum state $|\psi\rangle = [\frac{1}{\sqrt{2}}, \,\frac{-i}{\sqrt{2}} ]^T$, equivalently written as $|\psi\rangle = \frac{1}{\sqrt{2}} |0\rangle + \frac{-i}{\sqrt{2}} |1\rangle$, is a valid state vector in a two-dimensional Hilbert space. The basis states $|0\rangle$ and $|1\rangle$ each have a probability of $\left| \frac{1}{\sqrt{2}} \right|^2 = \frac{1}{2}$. Since $\left| \frac{1}{\sqrt{2}} \right|^2 + \left| \frac{-i}{\sqrt{2}} \right|^2 = \frac{1}{2} + \frac{1}{2} = 1$, the normalization condition is satisfied. The conjugate transpose, $\langle \psi | = [\frac{1}{\sqrt{2}} , \frac{i}{\sqrt{2}}]$, is a row vector, enabling the computation of the density matrix via the outer product $|\psi\rangle \langle \psi |$.

The general quantum analog of the classical belief state $\vec{x}_t$ is the density matrix $\hat{\rho}$, which captures both classical uncertainties and quantum indeterminacy. The diagonal elements of $\hat{\rho}$ represent the probabilities of being in each system state, satisfying the normalization condition $\operatorname{tr}(\hat{\rho}) = 1$. The off-diagonal elements encode quantum coherences and entanglement, phenomena with no classical counterpart. For example, the density matrix for the pure state $|\psi\rangle = \frac{1}{\sqrt{2}} |0\rangle + \frac{-i}{\sqrt{2}} |1\rangle$ is
$$\hat{\rho} = |\psi\rangle \langle \psi | = \begin{bmatrix} \frac{1}{\sqrt{2}}\vspace{0.1cm} \\ \frac{-i}{\sqrt{2}} \end{bmatrix} \begin{bmatrix} \frac{1}{\sqrt{2}}\; \frac{i}{\sqrt{2}} \end{bmatrix} = \begin{bmatrix} \frac{1}{2} & \frac{i}{2}\vspace{0.1cm} \\ \frac{-i}{2} & \frac{1}{2} \end{bmatrix},$$
where the diagonal entries $\frac{1}{2}$ and $\frac{1}{2}$ confirm equal probabilities for states $|0\rangle$ and $|1\rangle$, and the off-diagonal terms $\frac{i}{2}$ and $ \frac{-i}{2}$ reflect quantum coherence between the states.

In the classical setting, the joint distribution of two systems is represented by the tensor product $\vec{x}_1 \otimes \vec{x}_2$. In the quantum setting, the analogous concept is the multi-particle density matrix $\hat{\rho}_{1} \otimes \hat{\rho}_{2}$, formed by the tensor product of two density matrices $\hat{\rho}_{1}$ and $\hat{\rho}_{2}$ corresponding to quantum systems $1$ and $2$. As a valid density matrix, $\hat{\rho}_{1 2}= \hat{\rho}_{1} \otimes \hat{\rho}_{2}$ satisfies $\operatorname{tr}(\hat{\rho}_{1 2}) = 1$, with its diagonal elements representing the probabilities of the joint states in the Cartesian product of the basis states of $1$ and $2$. This joint density matrix serves as the quantum counterpart to the classical joint probability distribution, but its off-diagonal elements encoding additional quantum information, such as coherences and entanglement, has no classical equivalent. From the joint distribution, analogous to the classical marginalization $\vec{x}_1 = \sum_{x_2} \vec{x}_1 \otimes \vec{x}_2$, we can extract the state of one particle in a quantum system using the partial trace operation. For a two-particle system described by the joint density matrix $\hat{\rho}_{1 2}$, the partial trace over the particle $2$ yields the reduced density matrix for the particle $1$ as
$$\hat{\rho}_{1} = \operatorname{tr}_{2}(\hat{\rho}_{1 2}) = \sum_j {_2}\langle j| \hat{\rho}_{1 2} |j\rangle_{2},$$ 
where $|j\rangle_{2}$ denotes the vector with 1 at the $j$th coordinate of the particle $2$ and 0 everywhere else.
This operation sums over the diagonal elements of the basis states of the particle $2$ to be disregarded, producing the reduced density matrix $\hat{\rho}_{1}$ for the subsystem of interest.

In classical probability, conditional probability quantifies the likelihood of an event $\vec{x}$ given an observation $y$, defined as $P(\vec{x}|y) = \frac{P(y,\vec{x})}{P(y)}$, where $P(y,\vec{x})$ is the joint probability of $\vec{x}$ and $y$, and the normalizing factor $P(y)$ is the marginal probability of $y$. In the quantum setting, the analogous concept for computing the conditional state of a system given an observation $y$ involves projection and partial trace operations, expressed as 
$$P(\text{states} \,| y) = \operatorname{tr}_Y(\hat{P}_y \hat{\rho}_{XY} \hat{P}_y^\dagger),$$ 
where $\hat{\rho}_{XY}$ is the joint density matrix of the system and observation, $\hat{P}_y$ is the von Neumann projection operator onto the subspace corresponding to observation $y$, the $\dagger$ symbol denotes the Hermitian conjugate, and $\operatorname{tr}_Y$ traces out the observation subsystem to yield the reduced state of the system. Von Neumann projection operators are Hermitian, idempotent operators ($\hat{P}_y = \hat{P}_y^\dagger$, $\hat{P}_y^2 = \hat{P}_y$) that project a quantum state onto a specific subspace associated with a measurement outcome, such as $y$, effectively collapsing the state to reflect that outcome while preserving quantum coherence within the subspace. This process generalizes classical conditioning by incorporating quantum measurement and entanglement, with the projection ensuring the state is updated according to the observed outcome.

\subsection{Hidden Quantum Markov Models}
\label{sec:HQMM}
HQMMs extend classical HMMs into the quantum domain by parameterizing state transitions through a set of Kraus operators. These operators govern the evolution of quantum states, allowing HQMMs to capture richer dependencies and correlations that go beyond classical probabilistic structures. Kraus operators, denoted ${ {K}_i }$, are linear operators that describe a general quantum operation acting on a quantum system’s density matrix ${\rho}$. Such an operation transforms an initial density matrix ${\rho}$ into a new density matrix ${\rho}'$ via: ${\rho}' = \sum_i {K}_i {\rho} {K}_i^\dagger$. This form is known as the operator-sum representation of a quantum operation $\mathcal{K}$. Every completely positive (CP) map admits this representation:
$$\mathcal{K}(\rho) = \sum_i {K}_i \rho {K}_i^\dagger,$$
and conversely, any map of this form is CP. A quantum operation $\mathcal{K}$ is a CP trace non-increasing linear map on the space of density operators. 

Let’s first provide an example of Kraus operators to illustrate its functionality and properties. 
Consider a single qubit undergoing a dephasing process, where quantum coherence (off-diagonal elements of the density matrix) is reduced with probability $p$. The Kraus operators for a dephasing channel are
$$K_0 = \sqrt{1 - \frac{p}{2}} \mathbb{I} \quad\text{and}\quad K_1 = \sqrt{\frac{p}{2}} \hat{\sigma}_z,$$
where $\mathbb{I} = \begin{bmatrix} 1 & 0 \\ 0 & 1 \end{bmatrix}$ is the identity matrix, and $\hat{\sigma}_z = \begin{bmatrix} 1 & 0 \\ 0 & -1 \end{bmatrix}$ is the Pauli-Z operator. The action on a density matrix $\rho = \begin{bmatrix} \rho_{00} & \rho_{01} \\ \rho_{10} & \rho_{11} \end{bmatrix}$ is
$$\rho' = K_0 \rho K_0^\dagger + K_1\rho K_1^\dagger = \begin{bmatrix} \rho_{00} & (1 - p) \rho_{01} \\ (1 - p) \rho_{10} & \rho_{11} \end{bmatrix}.$$
The off-diagonal elements (coherences) are scaled by $1 - p$, reflecting dephasing, and the completeness relation holds:
$$K_0^\dagger K_0 + K_1^\dagger K_1 = \left(1 - \frac{p}{2}\right) \mathbb{I} + \frac{p}{2} \mathbb{I} = \mathbb{I}.$$

The trace non-increasing condition is expressed as:
$$\sum_i {K}_i^\dagger {K}_i \leq \mathbb{I},$$
where $\mathbb{I}$ is the identity operator. Quantum operations are classified as trace-preserving (TP), satisfying $\sum_i {K}_i^\dagger {K}_i = \mathbb{I}$, or trace-decreasing, where $\sum_i {K}_i^\dagger {K}_i < \mathbb{I}$. TP operations, often called quantum channels or CPTP maps, map density operators to density operators. If quantum operations are trace-decreasing, they are typically part of a larger structure, i.e., a complete set of trace-decreasing operations ${ \mathcal{K}_s }$, which together form a stochastic quantum operation.
The sum $\mathcal{K} = \sum_s \mathcal{K}_s$ is TP, ensuring the final state is a properly normalized density operator when the generated symbol is discarded. Stochastic quantum operations encompass a range of processes, from unitary transformations ($ \mathcal{K}_s(\cdot) = {U} \cdot {U}^\dagger $) to von Neumann measurements ($ \mathcal{K}_s(\cdot) = {P}_s \cdot {P}_s $), as well as generalized measurements such as positive operator-valued measures (POVMs).

Stochastic quantum operations represent quantum measurements associated with a specific outcome $s$. The probability of successfully applying a given operation $\mathcal{K}_s$ to a quantum state $\rho$ is
$$P(s) = \operatorname{tr}[\mathcal{K}_s (\rho)].$$
We can associate a set of quantum operations $\{ \mathcal{K}_s \}$ with a set of outcomes $ \{ s \}$, where each outcome $s$ occurs with probability $P(s)$, and the conditional state after observing outcome $s$ is given by
 $$\rho_s = \frac{\mathcal{K}_s (\rho)}{P(s)}.$$
Let $\{\hat{K}_i(s)\}_i$ denote the Kraus operators in the operator-sum representation of $\mathcal{K}_s$. The total probability summing to 1, i.e., $\sum_s P(s) = 1$, is ensured because the composite operation $\sum_s \mathcal{K}_s$ is trace-preserving, satisfying:
\begin{align}
	\label{eqn:TP}
\sum_{s,i} \hat{K}_i^\dagger(s) \hat{K}_i(s) = \mathbb{I}.
\end{align}
This demonstrates that the complete set of Kraus operators forms a quantum channel that corresponds to performing the measurement and discarding the outcome, yielding the final state
$$\tilde{\rho} = \sum_s P(s) \rho_s = \sum_{s,i} \hat{K}_i(s) \rho \hat{K}_i^\dagger(s).$$

Leveraging the framework of Kraus operators, HQMMs provide a quantum analog to classical HMMs.
\begin{definition}[\cite{monras2010hidden}]
	\label{def:HQMM}
	A HQMM is a quantum system $\rho$ together with a set of quantum operations $\mathcal K_s$ such that $\sum_s \mathcal K_s $ is trace-preserving. At every time step a symbol is generated with probability $P(s)=\operatorname{tr} [\mathcal K_s\rho]$ and the state vector is updated to $\rho_s=\mathcal K_s\rho/P(s)$.
\end{definition}

\subsection{HQMMs Generalize HMMs}
\label{sec:HQMM_HMM}
 \cite{monras2010hidden} demonstrates that HQMMs are the quantum analogs of classical HMMs, obtained by analogizing the classical Markov process and its transition matrices into a quantum process governed by stochastic quantum operations. The analogies includes: classical states being mapped to quantum states (described by density matrices); transition matrices $ T_s $ (defined in \eqref{eqn:T}) being substituted with quantum operations $ \mathcal{K}_s $ represented by Kraus operators $ \hat{K}_i(s) $ satisfying \eqref{eqn:TP}; probability calculations shifting from matrix products to quantum traces. 
However, their paper emphasizes a functional analogy, leaving the exact correspondence implicit within the transition from classical to quantum formalism. This naturally raises two questions for HMM users: 
\begin{itemize}
\item Do HQMMs generalize HMMs in the conventional sense?
\item If so, can one explicitly extract HMM information from an HQMM? 
\end{itemize}
Merely based on the definition of HQMM in Definition \ref{def:HQMM}, we are not able to give firm answers to these questions.

\begin{figure}
	\centering
	\mbox{\Qcircuit @C=1em @R=1em {
			\lstick{\hat{\rho}_{t-1}} & \multigate{1}{\hat{U}_{1}} & \qw & \qw &  &  & & \\
			\lstick{\hat{\rho}_{env}^{(1)}} & \ghost{\mathcal{F}} & \qw & \multigate{1}{\hat{U}_{2}} & \qw & \qw &  {\hat{\rho}_t} \\
			\lstick{\hat{\rho}_{env}^{(2)}} & \qw & \qw & \ghost{\mathcal{F}} & \measureD{}\\
	}}
	\caption{Full quantum circuit to implement HMM in equation \eqref{eqn:HMM_quantum}}
	\label{fig:HMM_quantum}
\end{figure}
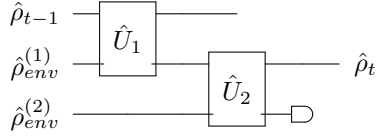

\cite{srinivasan2018learning} first explored simulating classical HMMs on quantum circuits using qudits (a $d$-state quantum system, akin to qubits or `quantum bits' which are 2-state quantum systems. Quantum operations must be represented by unitary matrices to preserve the 2-norm of the state. Since column-stochastic transition and emission matrices of classical HMMs are not generally unitary, they introduced environment particles (termed `ancilla particles' in their paper) to construct unitary operators. These environment particles are represented by two density matrices $\hat{\rho}_{\text{env}}^{(1)}$ and $\hat{\rho}_{\text{env}}^{(2)}$, both with one in the first entry and all other entries zero.
They defined two unitary operators, $\hat{U}_1$ and $\hat{U}_2$, derived from the transition matrix $\mathbf{A}$ and emission matrix $\mathbf{C}$, respectively. The operator $\hat{U}_1$ evolves the joint state $\hat{\rho}_{t-1} \otimes \hat{\rho}_{\text{env}}^{(1)}$ to perform the Markovian transition, while $\hat{U}_2$ updates the resulting state to encode probabilities of observable outputs. The quantum forward algorithm for simulating an HMM is expressed as:
 \begin{equation}
	\label{eqn:HMM_quantum}
\hat{\rho}_t \propto \text{Tr}_{\hat{\rho}_{\text{env}}^{(2)}} \left( \hat{P}_y \hat{U}_2 \left( \text{Tr}_{\hat{\rho}_{t-1}} \left( \hat{U}_1 (\hat{\rho}_{t-1} \otimes \hat{\rho}_{\text{env}}^{(1)}) \hat{U}_1^\dagger \right) \otimes \hat{\rho}_{\text{env}}^{(2)} \right) \hat{U}_2^\dagger \hat{P}_y \right).
\end{equation}
Here, $\hat{P}_y$ is a von Neumann projection operator corresponding to observation $y$. It selects the probability distribution entries matching the observation $y$, setting others to zero, analogous to indexing a joint probability distribution with an indicator vector.
The quantum circuit implementing Equation \eqref{eqn:HMM_quantum} is depicted in Figure \ref{fig:HMM_quantum}, with the D-shaped symbol representing $\hat{P}_y$.

The quantum implementation of HMMs described above requires further validation to clarify how it incorporates the classical HMM framework. In other words, if the quantum implementation generalizes the classical HMM, it remains to be shown how classical information can be retrieved from the quantum model. To bridge this gap, we prove the following proposition in Appendix \ref{proof:thm:diagonal_match}, rigorously revealing the implication in a general model setting with a hidden state dimension of $n=2$ and an observation dimension of $m=3$. This choice simplifies the demonstration, but the approach can be naturally extended to high dimensional cases with straightforward technical modifications.
\begin{proposition}
	\label{thm:diagonal_match}
Consider the HMM with general transition matrix $$
\mathbf{A} = \begin{pmatrix} a & 1-b \\ 1-a & b \end{pmatrix}, \qquad 0 \leq a, b \leq 1,
$$
and general emission matrix
$$
\mathbf{C} = \begin{pmatrix} e_0 & f_0 \\ e_1 & f_1 \\ e_2 & f_2 \end{pmatrix}, \qquad e_0 + e_1 + e_2 = 1, \qquad f_0 + f_1 + f_2 = 1.
$$
With suitably designed unitaries $\hat{U}_1$ and $\hat{U}_2$, the diagonal entries of the normalized density matrix generated by equation \eqref{eqn:HMM_quantum} coincide with the posterior distribution of the HMM:
$$
x_t(j | y) = \frac{P(y | j) (A x_{t-1})(j)}{\sum_j P(y | j) (A x_{t-1})(j)}, \qquad j \in \{0, 1\}.
$$
\end{proposition}

\cite{srinivasan2018learning} further showed that $\hat{\rho}_t$ in equation \eqref{eqn:HMM_quantum} can be simplified written as
 \begin{equation} 
 	\hat{\rho}_t = \frac{\hat{K}_{y_{t-1}}\left(\text{tr}_{\hat{\rho}_{t-1}}(\hat{U}_1(\hat{\rho}_{t-1} \otimes \hat{\rho}_{X_{t}})\hat{U}_1^\dagger)\right)\hat{K}_{y_{t-1}}^\dagger}{\text{tr}\left(\hat{K}_{y_{t-1}}\left(\text{tr}_{\hat{\rho}_{t-1}}(\hat{U}_1(\hat{\rho}_{t-1} \otimes \hat{\rho}_{X_{t}})\hat{U}_1^\dagger)\right)\hat{K}_{y_{t-1}}^\dagger\right)},
\end{equation}
where $\hat{K}_{y_t}$ is the Kraus operator that implements Bayesian quantum update after observation $y_t$. To simplify $\hat{U}_1$, instead of using a single Kraus operator, a set of $n$ Kraus operators $\{\hat{K}_w\}$ is employed. Post-multiplying each operator in $\{\hat{K}_w\}$ with ${\hat{K}_{y_t}}$ yields a new set of Kraus operators ${\hat{K}_{w_{y_t},y_t}}$, resulting in the expression:
\begin{equation}
	\hat{\rho}_t = \frac{\sum_{w_{y_t}} \hat{K}_{w_{y_t},y_{t}}\hat{\rho}_{t-1}\hat{K}_{w_{y_t},y_{t}}^\dagger}{\text{tr}\left(\sum_{w_{y_t}} \hat{K}_{w_{y_t},y_{t}}\hat{\rho}_{t-1}\hat{K}_{w_{y_t},y_{t}}^\dagger\right)}.
	\label{eq:hqmm}
\end{equation}
To comply with Definition \ref{def:HQMM}, the set of Kraus operators should be trace-preserving, satisfying:
$$\sum_{w_{y_t}} \hat{K}^\dagger_{w_{y_t},y_{t}} K_{w_{y_t},y_{t}} = \mathbb{I}.$$
Equation \ref{eq:hqmm} provides the forward algorithm for HQMMs. Its denominator represents the probability of emitting output $y$ given the previous state $\hat{\rho}_{t-1}$. The number of parameters in an HQMM is determined by the dimension of the latent state space 
$n$, the dimension of the measurement space 
$m$, and the number of Kraus operators $w$ associated with each measurement.

\subsection{Adversarially Corrupted HQMM}
\label{sec:AC_HQMM}
Learning the properties of quantum states is a central challenge in quantum computing, particularly in the NISQ era~\citep{preskill2018quantum}, where SPAM errors are prevalent. While SPAM noise models capture some device imperfections, they fail to account for broader disruptions such as environmental fluctuations, hardware instabilities, or even intentional interference. To address this limitation, \citet{aliakbarpour2025adversarially} introduced the adversarial corruption model, in which an adversary can arbitrarily corrupt a $\gamma$-fraction of measurement outcomes, demonstrating that this model generalizes SPAM noise and enables learning an unknown rank-$r$ quantum state to within $\tilde{O}(\gamma\sqrt{r})$ in trace distance. Extending this perspective to dynamic quantum systems, HQMMs operate in a more intricate setting that combines quantum graphical dependencies with sequential data modeling, making them vulnerable to both physical and adversarial perturbations. Developing robust HQMMs under such adversarial corruption is therefore crucial to ensure reliable inference and prediction in practical quantum computing applications.

We introduce the concept of $\gamma$-adversarial corruption to the HQMM literature by defining the AC-HQMM. Consider an initial quantum state $\rho_0 \in \mathbb{C}^{n \times n}$, where $\rho_0$ is a density matrix, and a set of quantum operations $\{\mathcal{K}_s\}_{s=1}^m$, each associated with a measurement outcome $s \in \{1, \dots, m\}$. Each quantum operation $\mathcal{K}_s$ is represented in the operator-sum form by a set of Kraus operators $\{\hat{K}_{s,q}\}_{q=1}^Q$, such that $$\mathcal{K}_s(\rho) = \sum_{q=1}^Q \hat{K}_{s,q} \rho \hat{K}_{s,q}^\dagger,$$ and the operations satisfy the trace-preserving condition:
\begin{equation}
	\label{eqn:TPC}
\sum_{s=1}^m \sum_{q=1}^Q \hat{K}_{s,q}^\dagger \hat{K}_{s,q} = \mathbb{I}.
\end{equation}

The observation matrix $ Y \in \mathbb{Z}^{N \times T} $ is generated to represent $ N $ sequences, each of length $ T $, where each element $ Y(j,i) \in \{1, \dots, m\} $ corresponds to a measurement outcome observed at time step $ i $ in sequence $ j $. The generation process proceeds as follows:

\begin{enumerate}
	\item \textbf{Initialization.} For each sequence $ j = 1, \dots, N $, initialize the quantum state as $\rho = \rho_0$.
	\item \textbf{Time Evolution.} For each time step $ i = 1, \dots, T $ in sequence $ j $:
	\begin{enumerate}
		\item \textbf{Probability Computation.} Compute the probability of each outcome $ s \in \{1, \dots, m\} $ according to:
		\[
		P(s) = \operatorname{tr}[\mathcal{K}_s (\rho)] = \operatorname{tr}\left[ \sum_{q=1}^Q \hat{K}_{s,q} \rho \hat{K}_{s,q}^\dagger \right].
		\]
		This defines a probability distribution $ \{ P(s) \}_{s=1}^m $ over the outcomes, where $\sum_{s=1}^m P(s) = 1$ due to the trace-preserving property of the operations.
		\item \textbf{Outcome Sampling.} Sample an outcome $ s \in \{1, \dots, m\} $ with probability $ P(s) $, and set $ Y(j,i) = s $.
		\item \textbf{State Update.} Update the quantum state conditioned on the observed outcome $ s = Y(j,i) $:
		\[
		\rho_s = \frac{\mathcal{K}_s (\rho)}{P(s)} = \frac{\sum_{q=1}^Q \hat{K}_{s,q} \rho \hat{K}_{s,q}^\dagger}{\operatorname{tr}\left[ \sum_{q=1}^Q \hat{K}_{s,q} \rho \hat{K}_{s,q}^\dagger \right]}.
		\]
		Set $\rho = \rho_s$ for the next time step.
	\end{enumerate}
	\item \textbf{Sequence Completion.} Repeat step 2 for all $ T $ time steps to generate the sequence $ Y(j, 1:T) $. Reset $\rho = \rho_0$ and repeat for all $ N $ sequences.
	\item \textbf{$\gamma$-adversarial corruption.} The resulting matrix $ Y $ contains $ N $ independent observation sequences, each of length $ T $, where each observation $ Y(j,i) $ is generated according to the HQMM dynamics. An adversary can arbitrarily corrupt a $N\gamma$ sequences out of these $ N $ observation sequences.
\end{enumerate}
We note that the adversary is typically assumed to have perfect knowledge of the measurements. Even though there are measurements $ Y(:,i) $ for each time step $ i \in \{1, \dots, T\} $, our definition allows the corruption to occur at any time step, requiring only that $ N\gamma $ out of these $ N $ observation sequences are corrupted eventually. Clearly, when $ \gamma = 0 $, the AC-HQMM reduces to the standard HQMM.

\section{Likelihood-based HQMM learning}
\label{LHL}
After properly defining AC-HQMM, this section aims to address the four questions raised in the Introduction. Specifically, we seek to design a corruption robust mechanism that is broadly adaptable to HQMM learning algorithms and remains reliable under adverse conditions. We also explore the possibility of developing quantum learning algorithms that are not only robust under arbitrary corruptions but also compatible with different metrics across diverse scenarios. Finally, beyond these extended capabilities, we aim to fundamentally improve existing algorithms so that they consistently produce valid quantum operators without approximation errors. To achieve these goals, we begin with the mainstream setting and examine the first, and so far only, known algorithm that guarantees valid quantum operators, as presented in Section \ref{sec:ILA}. We propose our robust learning algorithm for AC-HQMMs in Section \ref{sec:RILA}.

\subsection{Iterative Learning Algorithm}
\label{sec:ILA}
Likelihood-based approaches for learning the Kraus operators of an HQMM are the predominant method in the HQMM field. They aim to maximize the log-likelihood of an HQMM for an observed sequence $y_1, \ldots, y_T$, expressed as
\begin{equation} 
	\label{eqn:likelihood}
	\mathcal{L} = \ln \text{tr}\left(\sum_{w_{y_n}}\hat{K}_{w_{y_n},y_n}\ldots \left(\sum_{w_{y_1}}\hat{K}_{w_{y_1}, y_1}\hat{\rho}_0\hat{K}_{w_{y_1},y_1}^\dagger\right)\ldots \hat{K}_{w_{y_n},y_n}^\dagger\right).
\end{equation}
Directly maximizing the log-likelihood using gradient descent is challenging due to the constraints on Kraus operators and the risk of underflow issues with long sequences. \cite{srinivasan2018learning} proposed to learn an $nmw \times n$ matrix $\kappa^*$, which represents the set of $wm$ Kraus operators ${\hat{K}_{w_y, y}}$ for $y\in \{1,\ldots,m\}$, each of dimension $n \times n$, stacked vertically. 
Let $\kappa^*$ be the true matrix of stacked Kraus operators that maximizes the likelihood for the observed data. There exists a unitary operator $\hat{U}$ that transforms the current estimate $\kappa$ to $\kappa^*$, such that $\kappa^* = \hat{U}\kappa$. 

\begin{algorithm}[t!]
	\caption{ILA for HQMMs in \cite{srinivasan2018learning}}
	\begin{algorithmic}[1]
		\makeatletter
		\let\oldALC@it\ALC@it
		\renewcommand{\ALC@it}{\oldALC@it\vspace{0.1cm}}
		\makeatother
		\vspace{0.1cm}
		
		\item[] \hspace*{-\leftmargin}  \textbf{Input:} An $N \times T$ matrix $Y$, where $N$ is the number of data sequences and $T$ is the length of a stochastic sequence to be modeled.\vspace{0.1cm}
		\item[] \hspace*{-\leftmargin}  \textbf{Output:} A set of $wm$ Kraus operators $\{\hat{K}_{w_y, y}\}$ of dimension $n \times n$ that maximize the log-likelihood of the data, where $n$ is the dimension of the hidden state, $m$ is the number of outputs, and $w$ is the number of operators per output.\vspace{0.1cm}
		\STATE \textbf{Initialization:} Randomly generate a set of $wm$ Kraus operators ${\hat{K}_{w_y, y}}$ of dimension $n \times n$, and stack them vertically to form a matrix $\kappa$ of dimension $nmw \times n$. Define $b$ as the batch size, $B$ as the total number of batches, and $Y_b$ as a $b \times T$ matrix of randomly chosen data samples. Set $L$ as the number of iterations to modify $\kappa$ to maximize the likelihood of observing $Y_b$.\vspace{0.1cm}
		\FOR{batch = 1 to $B$}
		\STATE Randomly select $b$ sequences to process and construct matrix $Y_b$.
		\FOR{$l = 1$ to $L$}
		\STATE Randomly select rows $i$ and $j$ of $\kappa$ to modify, where $i < j$.
		\STATE Find $\theta = (\alpha, \phi, \psi, \delta)$ that maximizes the log-likelihood of $Y_b$ under the update:
		\vspace{-0.2cm}
		\[
		\begin{aligned}
			\kappa^i &\gets \left(e^{\frac{i\phi}{2}}e^{i\psi}\cos(\alpha)\right)\kappa^i + \left(e^{\frac{i\phi}{2}}e^{i\delta}\sin(\alpha)\right)\kappa^j \\
			\kappa^j &\gets \left(-e^{\frac{i\phi}{2}}e^{-i\delta}\sin(\alpha)\right)\kappa^i + \left(e^{\frac{i\phi}{2}}e^{-i\psi}\cos(\alpha)\right)\kappa^j
		\end{aligned}
		\]
		\vspace{-0.4cm}
		\ENDFOR
		\ENDFOR
	\end{algorithmic}
	\label{alg:ILA}
\end{algorithm}

The matrix $\hat{U}$ can be represented as a product of simpler matrices  in the form that
\begin{equation*}
	{\bf H}(i,j,\alpha,\phi,\psi,\delta) = \begin{bmatrix} 1 & \cdots & 0 & \cdots & 0 & \cdots & 0 \\ \vdots & \ddots & \vdots &  &  \vdots & & \vdots \\ 0 & \cdots & e^{\frac{i\phi}{2}}e^{i\psi}\cos\alpha & \cdots & e^{\frac{i\phi}{2}}e^{i\delta}\sin\alpha & \cdots & 0 \\ \vdots &  & \vdots & \ddots  &  \vdots & & \vdots \\ 0 & \cdots & -e^{\frac{i\phi}{2}}e^{-i\delta}\sin\alpha & \cdots & e^{\frac{i\phi}{2}}e^{-i\psi}\cos\alpha & \cdots & 0 \\ \vdots & & \vdots &  &  \vdots & \ddots & \vdots \\ 0 & \cdots & 0 & \cdots & 0 & \cdots & 1\end{bmatrix}.
\end{equation*}
Here, $i$ and $j$ denote the two rows in the matrix with non-trivial entries, while the parameters $\alpha$, $\phi$, $\psi$, and $\delta$ are angles that parameterize these entries. \cite{srinivasan2018learning} transfered the problem from learning the Kraus operators to identifying the sequence of $\{{\bf H}\}$ matrices that can take $\kappa$ to $\kappa^*$. Since the optimization problem is non-convex and the matrices $\{\bf H\}$ do not necessarily commute, their algorithm is not guaranteed to find the global maximum. Instead, it seeks a local maximum $\kappa^*$ by iteratively multiplying $\{\bf H\}$ matrices in a manner that increases the log-likelihood. In Algorithm~\ref{alg:ILA}, we present the ILA for HQMMs as proposed in~\cite{srinivasan2018learning}.

\subsection{Robust Iterative Learning Algorithm}
\label{sec:RILA}

\begin{algorithm}[t!]
	\caption{Remove Corrupted Rows by Entropy Filtering (RCR-EF)}
	\begin{algorithmic}[1]
		\makeatletter
		\let\oldALC@it\ALC@it
		\renewcommand{\ALC@it}{\oldALC@it\vspace{0.1cm}}
		\makeatother
		\vspace{0.1cm}
		
		\item[] \hspace*{-\leftmargin} \textbf{Input:} An $N \times T$ matrix $Y$, where $N$ is the number of sequences and $T$ is the length of each sequence; Number of corrupted sequences to filter $C$. \vspace{0.1cm}
		
		\item[] \hspace*{-\leftmargin} \textbf{Output:} A filtered matrix $Y$ containing the $N - C$ least anomalous sequences based on entropy, unique value counts, mean, and variance. \vspace{0.1cm}
		
		\IF{$C > 0$ and $C < N$}
		\STATE Initialize arrays $E$, $U$, $M$, and $V$ of size $N$ to store entropy, unique value counts, mean, and variance for each sequence.
		\FOR{$i = 1$ to $N$}
		\STATE Extract sequence $\eta = Y(i, :)$.
		\STATE Compute frequency counts $f$ for values in $\eta$ using histogram bins.
		\STATE Compute probabilities $p = f / T$, excluding zero probabilities.
		\STATE Compute Shannon entropy $E(i) = -\sum (p \cdot \log_2(p))$.
		\STATE Compute unique value count $U(i) = |\text{unique}(\eta)|$.
		\STATE Compute mean $M(i) = \frac{1}{T} \sum_{j=1}^T \eta(j)$.
		\STATE Compute variance $V(i) = \frac{1}{T-1} \sum_{j=1}^T (\eta(j) - M(i))^2$.
		\ENDFOR
		\STATE Compute z-scores: 
		\STATE \quad $Z_E = (E - \text{mean}(E)) / \text{std}(E)$,
		\STATE \quad $Z_U = (U - \text{mean}(U)) / \text{std}(U)$,
		\STATE \quad $Z_M = (|M - \text{mean}(M)|) / \text{std}(M)$,
		\STATE \quad $Z_V = (|V - \text{mean}(V)|) / \text{std}(V)$.
		\STATE Compute outlier scores: 
		$$S = -Z_E - Z_U + w_M \cdot Z_M + w_V \cdot Z_V,$$ where $w_M = 0.5$ and $w_V = 0.5$.
		\STATE Sort sequences by descending $S$ to obtain indices $I$.
		\STATE Select the top $N - C$ indices from $I$ to form $I_{\text{keep}}$.
		\STATE Set $Y = Y(I_{\text{keep}}, :)$.
		\ELSE
		\STATE Return $Y$ unchanged.
		\ENDIF
	\end{algorithmic}
	\label{alg:filter_corrupted_combined}
\end{algorithm}

In this paper, we propose the Robust Iterative Learning Algorithm (RILA) for AC-HQMMs. RILA, as the name suggests, is a robust version of ILA. We build on the ILA framework because it is not only the first—and so far only—known algorithm that guarantees valid quantum operators, but also stands out among its approximate alternatives due to its lack of hyperparameters to tune. These two features eliminate the need for restrictive conditions on matrix ranks of algorithm building blocks, as well as avoiding unstable performance associated with hyperparameter tuning.

In Algorithm~\ref{alg:filter_corrupted_combined}, we first present the RCR-EF algorithm, a robust preprocessing mechanism designed to mitigate adversarial or noise-induced corruption in spatiotemporal training data. Unlike standard adversarial robustness frameworks~\citep{aliakbarpour2025adversarially} and related works, which rely on the i.i.d. assumption across samples, our setting involves $N$ interdependent sequences of length $T$, where temporal dynamics and cross-sequence correlations violate independence. This structural dependence renders classical i.i.d.-based filtering and robust estimation techniques inapplicable. We model the input as an $N \times T$ matrix $Y$, where each row represents a stochastic sequence to be modeled, and up to $C$ rows may be corrupted by structured noise, adversarial perturbations, or sensor artifacts, threats that can severely bias parameter estimation or degrade predictive accuracy. To counteract this, RCR-EF evaluates each sequence across four statistically grounded anomaly indicators:  
Shannon entropy ($E_i$),  
unique value count ($U_i$),  
absolute mean deviation ($|M_i - \bar{M}|$), and  
variance ($V_i$), for each $i \in \{1,\ldots,N\}$.  
Each metric is standardized via z-scoring ($Z_E, Z_U, Z_M, Z_V$) to ensure comparability across heterogeneous signal distributions. A composite outlier score is then formed as  
\[
S_i = -Z_{E,i} - Z_{U,i} + 0.5 Z_{M,i} + 0.5 Z_{V,i},
\]  
where negative weights on entropy and uniqueness penalize erratic or overly repetitive patterns, while moderated positive weights highlight sequences with extreme location or scale. The $N - C$ sequences with the lowest $S_i$ are retained, producing a filtered matrix $Y$ enriched with coherent, informative signals.

\begin{algorithm}[t!]
	\caption{RILA for AC-HQMMs}
	\begin{algorithmic}[1]
		\vspace{0.1cm}
		
		\item[] \hspace*{-\leftmargin} \textbf{Input:} An $N \times T$-dimensional training data matrix $Y$;  Validation data matrix $Y_{\text{val}}$;
		Initial set of $wm$ Kraus operators $K_{\text{init}} = \{\hat{K}_{w_y, y}\}$ of dimension $n \times n$;
		Initial density matrix $\rho$;
		Batch size $b$;
		Number of batches $B$;
		Number of iterations per batch $I$;
		Number of resampling propsals $P$;
		Number of corrupted sequences to filter $C$; Optimization solver $S$.\vspace{0.1cm}
		
		\item[] \hspace*{-\leftmargin}  \textbf{Output:} A set of $ws$ Kraus operators $K_{\text{best}}=\{\hat{K}_{w_y, y}\}$ of dimension $n \times n$ that maximize the log-likelihood of the data.\vspace{0.1cm}
		\STATE \textbf{Initialization:} Stack $K_{\text{init}}$ vertically to form a matrix $\kappa$ of dimension $nmw \times n$. Set $L_{\text{best}}=-\infty$. \vspace{0.1cm}
		
		\STATE \textbf{Filtering:}  $Y_{\text{clean}} \leftarrow \text{RCR-EF}(Y,C)$ by calling Algorithm \ref{alg:filter_corrupted_combined}.
		\vspace{0.1cm}
		\FOR{batch = 1 to $B$}
		\STATE Randomly select $b$ sequences from $Y_{\text{clean}} $ to construct matrix $Y_b$.
		\FOR{iteration = 1 to $I$}
		\FOR{$p = 1$ to $P$}
		\STATE Randomly select rows $i$ and $j$ of $\kappa$, where $i < j$.
		\STATE Compute $L \gets f(Y_b, \kappa, \rho)$.
		\STATE Identify parameters $ \theta = (\alpha, \phi, \psi, \delta) $ that maximize $ f(Y_b, \kappa, \rho) $ using the solver $ S $, and generate $ \widetilde{\kappa}$ accordingly via $ \text{update}(\widetilde{\kappa}; \theta, i, j) $:
		\vspace{-0.2cm}
		\[
		\begin{aligned}
			\widetilde{\kappa}^i &\gets \left(e^{\frac{i\phi}{2}}e^{i\psi}\cos(\alpha)\right)\kappa^i + \left(e^{\frac{i\phi}{2}}e^{i\delta}\sin(\alpha)\right)\kappa^j ,\\
			\widetilde{\kappa}^j &\gets \left(-e^{\frac{i\phi}{2}}e^{-i\delta}\sin(\alpha)\right)\kappa^i + \left(e^{\frac{i\phi}{2}}e^{-i\psi}\cos(\alpha)\right)\kappa^j.
		\end{aligned}
		\]
		\vspace{-0.4cm}
		\STATE Compute updated log-likelihood $\widetilde{L} \gets f(Y_b, \text{update}(\widetilde{\kappa}; \theta, i, j), \rho)$.
		\IF{$\widetilde{L} > L$}
		\STATE Set $L \gets \widetilde{L}$ and $\kappa \gets \widetilde{\kappa}$.
		\ENDIF
		\STATE Store log-likelihood $L_p \gets L$ and Kraus operators $\kappa_p \gets \kappa$.
		\ENDFOR
		\STATE Compute weights,  for $p = 1, \dots, P$,
		\vspace{-0.2cm}
		$$w_p \gets \exp(-L_p - \min(-L_{1:P})) \bigg/ \sum_{j=1}^P \exp(-L_j - \min(-L_{1:P})).$$
		\vspace{-0.4cm}
		\STATE Sample an index $p^*$ from $\{1, \dots, P\}$ with probabilities $\{w_p\}$.
		\STATE Set $\kappa \gets \kappa_{p^*}$.
		\ENDFOR
		\STATE Compute validation log-likelihood $H(\text{batch}) \gets f(Y_{\text{val}}, \kappa, \rho)$
		\IF{$H(\text{batch}) > L_{\text{best}}$}
		\STATE Update $L_{\text{best}} \gets H(\text{batch})$ and $\kappa_{\text{best}} \gets \kappa$
		\ENDIF
		\STATE Transform $\kappa_{\text{best}}$ back to the form of $K_{\text{best}}$.
		\ENDFOR
	\end{algorithmic}
	\label{alg:RILA}
\end{algorithm}

The complete RILA procedure is detailed in Algorithm~\ref{alg:RILA}. After applying the RCR-EF filter (Algorithm~\ref{alg:filter_corrupted_combined}) to remove the $C$ most corrupted sequences and obtain the filtered matrix, training proceeds in a batched and iterative fashion over $B$ batches. In each batch, $b$ sequences are randomly sampled without replacement from the filtered $Y$ to form the batch matrix $Y_b \in \mathbb{R}^{b \times T}$. Then, $I$ iterations of parameter refinement are performed on $Y_b$, where each iteration involves a proposal phase consisting of $P$ candidate updates to the Kraus operator matrix $\kappa$. For each proposal, two distinct rows $i < j$ of $\kappa$ are randomly selected, and a targeted modification is proposed to their entries while enforcing physical validity constraints such as complete positivity and trace preservation. The initial log-likelihood $ L $ is computed as $ f(Y_b, \kappa, \rho) $. Subsequently, parameters $ \theta = (\alpha, \phi, \psi, \delta) $ are optimized using solver $ S $ to maximize $ f(Y_b, \kappa, \rho) $, yielding a tentative update $ \widetilde{\kappa} $ through the function $ \text{update}(\widetilde{\kappa}; \theta, i, j) $, defined explicitly as:
$$\begin{aligned}
	\widetilde{\kappa}^i &\gets \left(e^{i\phi/2} e^{i\psi} \cos(\alpha)\right) \kappa^i + \left(e^{i\phi/2} e^{i\delta} \sin(\alpha)\right) \kappa^j, \\
	\widetilde{\kappa}^j &\gets \left(-e^{i\phi/2} e^{-i\delta} \sin(\alpha)\right) \kappa^i + \left(e^{i\phi/2} e^{-i\psi} \cos(\alpha)\right) \kappa^j.
\end{aligned}$$
The updated log-likelihood $ \widetilde{L} $ is then evaluated as $ f(Y_b, \text{update}(\kappa; \theta, i, j), \rho) $. If $ \widetilde{L} > L $, the improvement is accepted by setting $ L$ as $\widetilde{L} $ and $ \kappa$ as $\widetilde{\kappa} $. Finally, before the new $\kappa$ value is passed to the input of the next proposal iteration, $L$ and $\kappa$ are stored as $L_p$ and $\kappa_p$ (where $p \in \{1, \ldots, P\}$) for resampling purposes once the proposal iterations are complete.

The users can choose any optimization solver. The current implementation supports two options: \texttt{fmincon} and \texttt{patternsearch}, both included in MATLAB's Optimization Toolbox. The \texttt{fmincon} function is a constrained nonlinear multivariable optimization solver, minimizing an objective function subject to bounds, linear inequalities/equalities, and nonlinear constraints. It employs iterative algorithms like interior-point, sequential quadratic programming, or active-set methods, starting from an initial guess and supporting user-defined gradients or Hessians for enhanced efficiency. However, limitations arise from its focus on local optima, sensitive to initial points and not guaranteeing global minima, requirements for differentiable functions. The \texttt{patternsearch} function is a direct search algorithm for finding the minimum of an unconstrained or bounded objective function without requiring derivatives, making it ideal for nonsmooth, discontinuous, or noisy problems. It starts from an initial point and explores a pattern of points (e.g., mesh) around the current best, adapting the search direction and scale based on successful polls, with options for linear/nonlinear constraints via penalty methods or filters. Its strengths lie in robustness to non-differentiable objectives, handling real-world scenarios like experimental data fitting or black-box optimizations, global search tendencies through pattern expansion, parallel evaluation support for speed, and no need for gradient computations, reducing sensitivity to function smoothness. However, limitations include slower convergence compared to gradient-based methods like \texttt{fmincon} due to exhaustive polling.

We illustrate the use of \texttt{fmincon} for differentiable log-likelihood functions and \texttt{patternsearch} for non-differentiable or non-smooth cases. When the log-likelihood includes non-differentiable components, such as absolute value penalties, indicator functions, or piecewise-defined terms, derivatives may not exist everywhere.
In maximum likelihood estimation, the objective is to maximize the log-likelihood function
$
\ell(\theta) = \log L(\theta; \text{data}),
$
where $L(\theta; \text{data})$ is the likelihood of parameters $\theta$ given the data. Often, a penalty term is added to regularize the model, prevent overfitting, or enforce certain properties (e.g., sparsity). The modified objective with an absolute value penalty (L1 regularization) takes the form
\begin{equation}
	\label{eqn:LL_P}
f(\theta) = \ell(\theta) - \lambda \sum_{i=1}^d |\theta_i|,
\end{equation}
where $\lambda > 0$ controls its strength and $\theta = (\theta_1, \dots, \theta_d)$ is the parameter vector. This penalty is non-differentiable because the absolute value function $|\theta_i|$ has a kink at $\theta_i = 0$, where the derivative is undefined. Gradient-based methods such as \texttt{fmincon} may therefore struggle to handle such cases effectively, whereas derivative-free optimizers like \texttt{patternsearch} are well-suited for these objectives. 

Beyond these extended capabilities, we demonstrate that RILA can fundamentally enhance existing algorithms without increasing the number of parameters through additional Kraus operators, thereby preserving computational complexity and mitigating overfitting. We introduce a sequential resampling step to further improve optimization efficiency. Specifically, after the iterative proposal phase, the log-likelihoods of the proposed solutions are transformed into normalized weights via an exponential function, where each weight is proportional to the exponentiated log-likelihood and adjusted by the maximum log-likelihood for numerical stability. A single proposal is then resampled from the $P$ candidates according to these weights, giving higher-weighted proposals a greater chance of selection. The Kraus operators associated with the selected proposal are carried forward to the next iteration. After $I$ iterations, the selected Kraus operators are evaluated on a validation dataset, and their log-likelihood is monitored. If the validation log-likelihood surpasses the previous best value, the corresponding Kraus operators are saved as the current best solution. This procedure repeats for a specified number of batch iterations, enabling efficient exploration of the parameter space while converging toward an optimal set of Kraus operators that maximize the log-likelihood function.

The sequential resampling procedure enhances the ability of local optimization methods, \texttt{fmincon} and \texttt{patternsearch}, despite their tendency to converge to local minima near the initial starting point. These local optimizers often become trapped in suboptimal basins of attraction, lacking the mechanism to explore the broader parameter space effectively. While global optimizers could address this limitation, they are often computationally prohibitive in this context. Instead, the resampling procedure introduces stochasticity and diversity by generating $ P $ proposals per iteration, each involving a random selection of operators followed by local optimization from a consistent initial condition (e.g., $\theta = [0; 0; 0; 0]$). These proposals sample diverse perturbations to the current solution. By assigning weights based on the exponential of the log-likelihoods, favoring proposals with better objective values, and probabilistically selecting one proposal, the algorithm balances exploitation of promising regions with exploration of the parameter space. Over multiple iterations, this iterative refinement increases the likelihood of escaping poor local minima and converging toward a globally better solution compared to a single deterministic local search.

\section{Numerical Anlaysis}
\label{NA}
In this section, we present a comprehensive evaluation of RILA against the baseline ILA and the classical EM algorithm across three benchmark HQMM learning tasks. We begin with the 2-state 4-output HQMM (denoted as (2,4)-HQMM) from \cite{monras2010hidden} in Section~\ref{sec:HQMM1}, followed by the (2,6)-HQMM (denoted as (2,6)-HQMM) introduced in \cite{srinivasan2018learning} in Section~\ref{sec:HQMM2}, and conclude with a classical 8-state 8-output HMM (denoted as (8,8)-HMM) in Section~\ref{sec:HMM3}. Beyond replicating the standard clean-data experiments reported in the original works, we systematically assess robustness by introducing adversarial corruption across all three models. To comprehensively assess performance, we evaluate each method using two likelihood objectives: the standard (regular) log-likelihood and a penalized (irregular) variant that incorporates regularization terms to enforce physical constraints and mitigate overfitting.
In all experiments, we use $N = 30$ training sequences and 5 validation sequences, each of length $T = 100$. We evaluate the algorithm under two batching configurations: $B = 4$ and $B = 8$ batches, with a fixed batch size of $b = 5$ sequences per batch. For each batch, we perform $I = 6$ iterations, and within each iteration, we generate $P = 10$ proposals for updating the Kraus operators.

\subsection{HQMM Example in \cite{monras2010hidden}}
\label{sec:HQMM1}
\cite{monras2010hidden} demonstrated that HQMMs offer a powerful framework for modeling stochastic processes generated by various quantum systems. They showed that the quantum formalism enables a more efficient representation of certain stochastic processes compared to classical models. In Section 6 of their work, they provided evidence that some stochastic processes can be described more compactly using HQMMs. Specifically, they analyzed a classical 4-state HMM and proved that it could not be reduced to fewer than three states while preserving the same stochastic behavior. In contrast, they constructed a 2-state HQMM that generates an equivalent process, highlighting the efficiency of the quantum approach.
\begin{figure}[t!]
	\centering
	\includegraphics[width=0.55\textwidth]{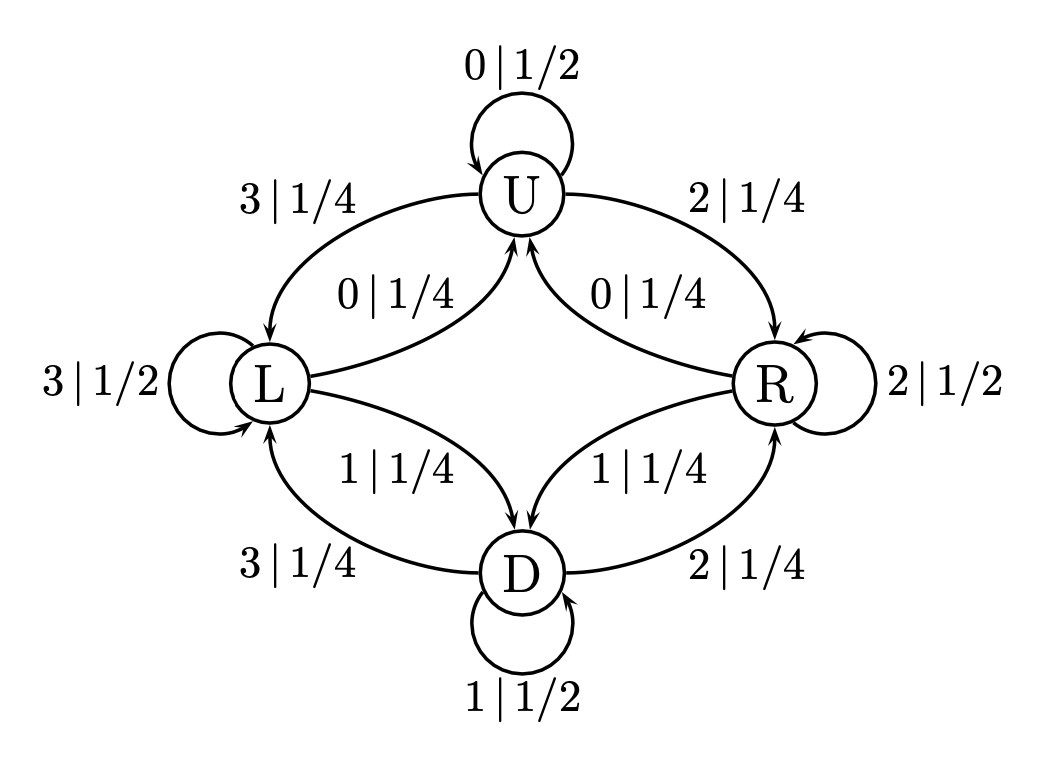}
	\caption{Diagram illustrating the stochastic generator for the HQMM as defined in Equation (\ref{eq:transfer_matrix}). Source: Figure 5 in \cite{monras2010hidden}.}
	\label{fig:diag}
\end{figure}

Consider the 4-symbol stochastic process with symbols $\{1, 2, 3, 4\}$, corresponding to a system with states labeled $U$, $D$, $R$, and $L$. The transition dynamics of this process are illustrated in Figure~\ref{fig:diag}, which depicts the stochastic generator as a directed graph with labeled transitions (e.g., probabilities such as $1/4$ and $1/2$ for transitions between states $U$, $D$, $R$, and $L$). The observation-specific transition operators defined in equation \eqref{eqn:T}, are given by
\begin{eqnarray}\label{eq:transfer_matrix}
	T_1=\left({\small
		\begin{array}{cccc}
			1/2 & 0 & 1/4 & 1/4\\
			0   & 0 & 0 & 0\\
			0 & 0 & 0 & 0\\
			0 & 0 & 0 & 0
	\end{array}}\right),
	\quad
	&T_2=\left({\small
		\begin{array}{cccc}
			0 & 0 & 0 & 0 \\
			0 & 1/2 & 1/4 & 1/4\\
			0 & 0 & 0 & 0\\
			0 & 0 & 0 & 0
	\end{array}}
	\right) \, , \nonumber \\
	T_3=\left({\small
		\begin{array}{cccc}
			0 & 0 & 0 & 0 \\
			0 & 0 & 0 & 0\\
			1/4 & 1/4 & 1/2 & 0\\
			0 & 0 & 0 & 0
	\end{array}}
	\right),
	\quad
	&T_4=\left({\small
		\begin{array}{cccc}
			0 & 0 & 0 & 0 \\
			0 & 0 & 0 & 0\\
			0 & 0 & 0 & 0\\
			1/4 & 1/4 & 0 & 1/2
	\end{array}}
	\right).
\end{eqnarray}

\cite{monras2010hidden} showed that the model described in equation \eqref{eq:transfer_matrix} can be generated by the following Kraus operators of a (2,4)-HQMM:
\begin{alignat*}{3}
	\hat{K}_{1,1} &= \begin{pmatrix} \frac1{\sqrt{2}} & 0 \vspace{0.03cm}\\ 0 & 0\end{pmatrix}, \hspace{1cm}
	&&\hat{K}_{2,1} = \begin{pmatrix} 0 & 0 \vspace{0.03cm}\\ 0 & \frac1{\sqrt{2}}\end{pmatrix}, \\
	\hat{K}_{3,1} &= \begin{pmatrix} \frac1{2\sqrt{2}} & \frac1{2\sqrt{2}} \vspace{0.09cm}\\ \frac1{2\sqrt{2}} & \frac1{2\sqrt{2}}\end{pmatrix}, \hspace{1cm}
	&&\hat{K}_{4,1} = \begin{pmatrix} \frac1{2\sqrt{2}} & -\frac1{2\sqrt{2}} \vspace{0.09cm}\\ -\frac1{2\sqrt{2}} & \frac1{2\sqrt{2}}\end{pmatrix}. 
\end{alignat*}
These operators satisfy the completeness relation for a quantum measurement \eqref{eqn:TPC} with $m=4$ and $Q=1$. 
In this HQMM, state evolution is unitary because each observable is associated with a single Kraus operator. We suppose this 2-state HQMM ($d=2$) starts with the initial density matrix $\rho_{0} = \begin{pmatrix} 1 & 0 \\ 0 & 0\end{pmatrix}$.

\begin{figure}[t!]
	\centering
	\includegraphics[width=0.73\textwidth]{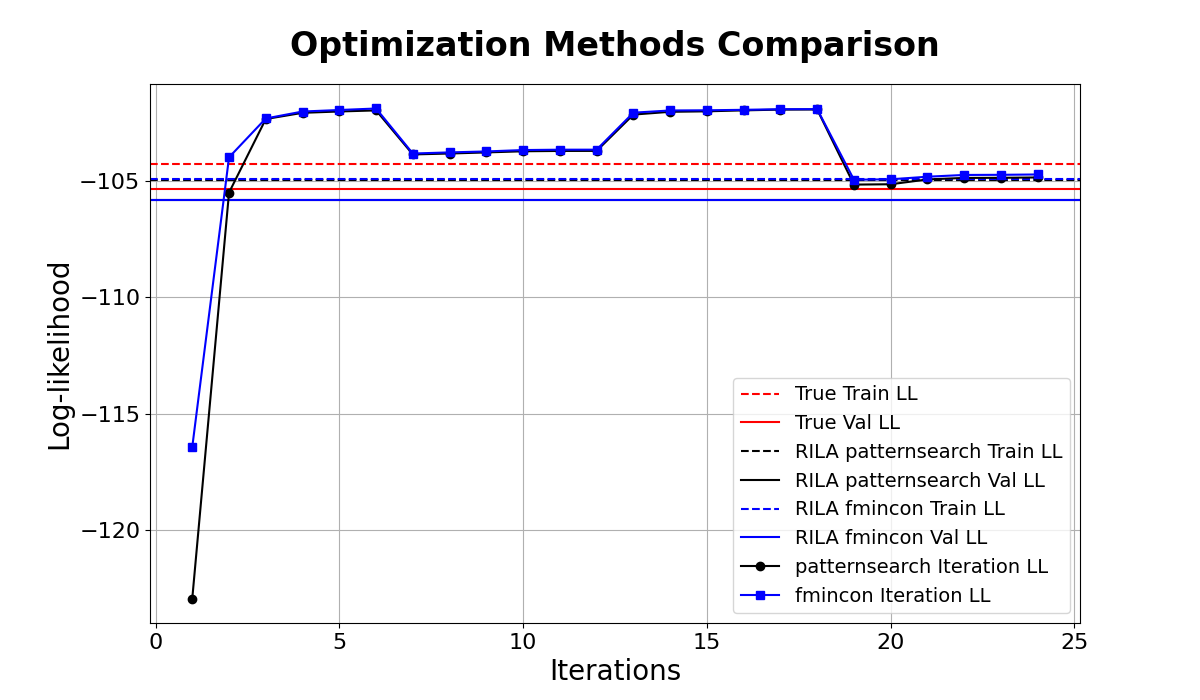}
	\caption{Comparison of HQMM learning performance using \texttt{patternsearch} and \texttt{fmincon} optimization methods on a dataset generated from the (2,4)-HQMM. 
		The plot shows true training (red dashed) and validation (red solid) log-likelihoods, alongside RILA learned values for \texttt{patternsearch} (black dashed for training, solid for validation) and \texttt{fmincon} (blue dashed for training, solid for validation). Iteration log-likelihoods are depicted with black circles (\texttt{patternsearch}) and blue squares (\texttt{fmincon}) across four batches.}
	\label{fig:pattersearch_fmincon}
\end{figure}

\begin{figure}[t!]
	\centering
	\includegraphics[width=0.73\textwidth]{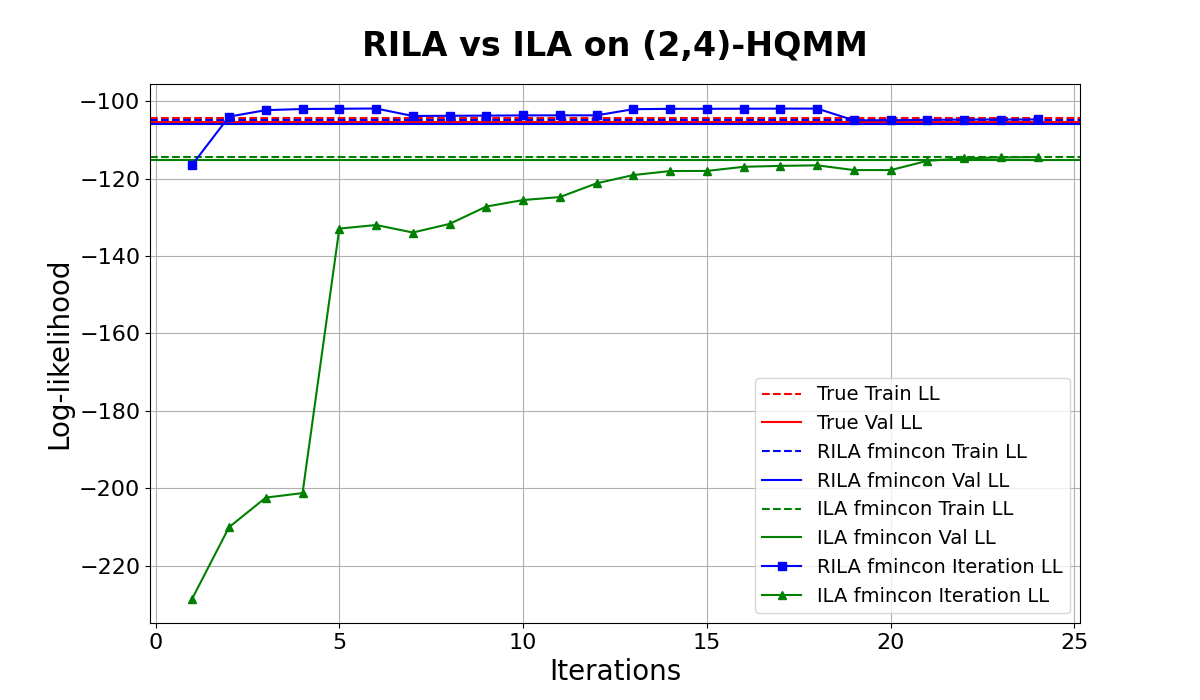}
	\caption{Comparison of HQMM learning performance using RILA and ILA across four batches on a dataset generated from the (2,4)-HQMM. The plot shows true training (red dashed) and validation (red solid) log-likelihoods, alongside RILA  learned values (blue dashed for training, solid for validation) and ILA  learned values (green dashed for training, solid for validation). Iteration log-likelihoods are depicted with blue squares (RILA) and green triangles (ILA) across four batches.}	\label{fig:HQMM_2_4_iterations}
\end{figure}

\begin{figure}[t!]
	\centering
	\includegraphics[width=0.73\textwidth]{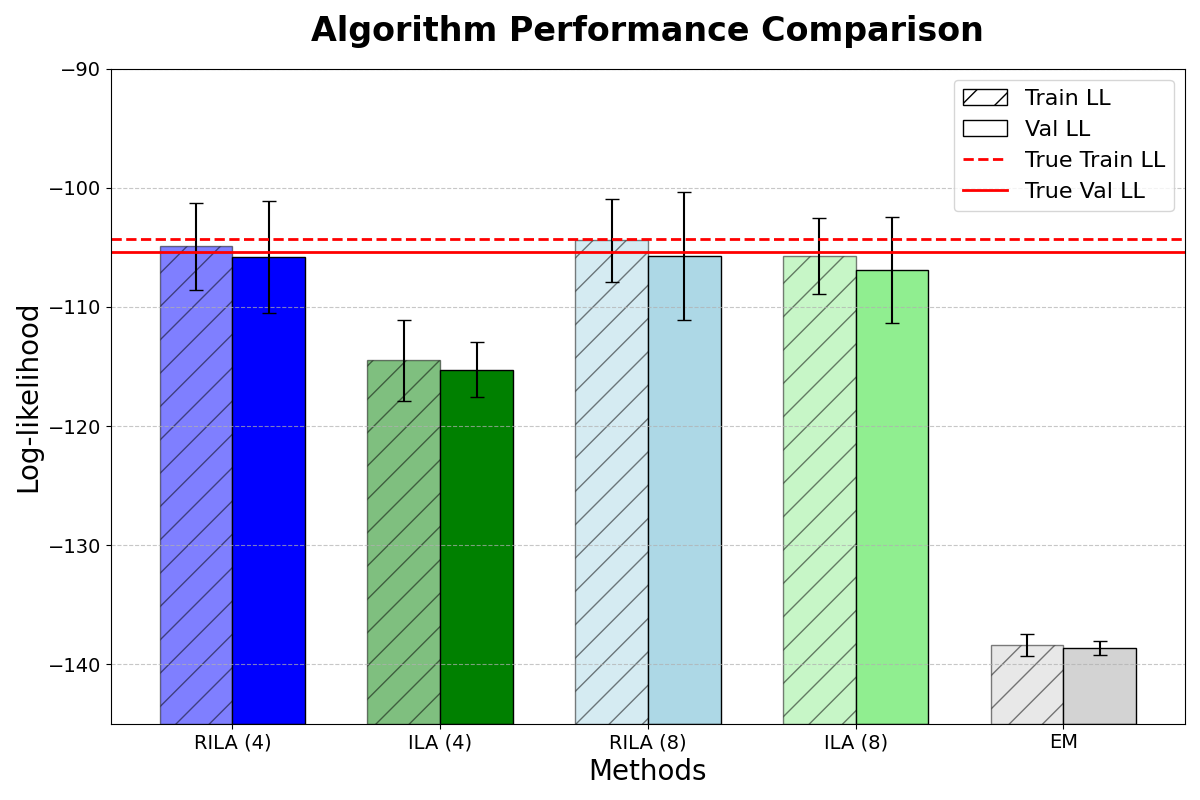}
	\caption{Grouped bar plot comparing the performance of RILA (4 and 8 batches), ILA (4 and 8 batches), and EM on a dataset generated from the (2,4)-HQMM. Each bar represents the mean log-likelihood, with training (dashed-pattern, lighter bars) and validation (solid, darker bars) values shown side by side for each method. Error bars indicate standard deviations. True training (red dashed) and validation (red solid) log-likelihoods are overlaid as horizontal reference lines. }	\label{fig:Algorithm_Performance_Comparison}
\end{figure}

Figure~\ref{fig:pattersearch_fmincon} presents a comparison of the learning performance of the HQMM using two optimization methods, \texttt{patternsearch} and \texttt{fmincon},  on a dataset generated from this (2,4)-HQMM. The plot displays the true training and validation log-likelihoods, along with the learned log-likelihood values obtained using RILA with both methods. It also tracks the iteration-wise log-likelihoods for each method across four batches. The results show that both optimizers achieve substantial improvement in the early iterations and stabilize after about five iterations, indicating convergence. Overall, both methods converge rapidly and exhibit nearly identical performance in this example. Figure \ref{fig:HQMM_2_4_iterations} compares the learning performance of RILA and ILA across four batches. The plot displays the true training and validation log-likelihoods together with the learned values obtained by RILA and ILA. The iteration log-likelihoods for RILA and ILA, illustrate the optimization trajectories. Both algorithms exhibit substantial improvement in the early iterations, with RILA converging within approximately five iterations and ILA requiring much more iterations to stabilize. RILA exhibits much faster convergence and achieves log-likelihoods closer to the true levels in this example. Figure \ref{fig:Algorithm_Performance_Comparison} presents a grouped bar plot comparing the performance of RILA, ILA, and EM. The bars show the mean training and validation log-likelihoods obtained using four and eight batches, with error bars indicating standard deviations. Both RILA and ILA markedly outperform the EM baseline, achieving log-likelihoods much closer to the true values. Between the two learning algorithms, RILA consistently attains higher log-likelihoods than ILA under both batch settings, demonstrating more effective optimization. The results further suggest that increasing the number of batches enhances stability, particularly for ILA, whereas RILA already attains accurate and stable performance with only four batches.

\begin{figure}[t!]
	\centering
	\includegraphics[width=0.69\textwidth]{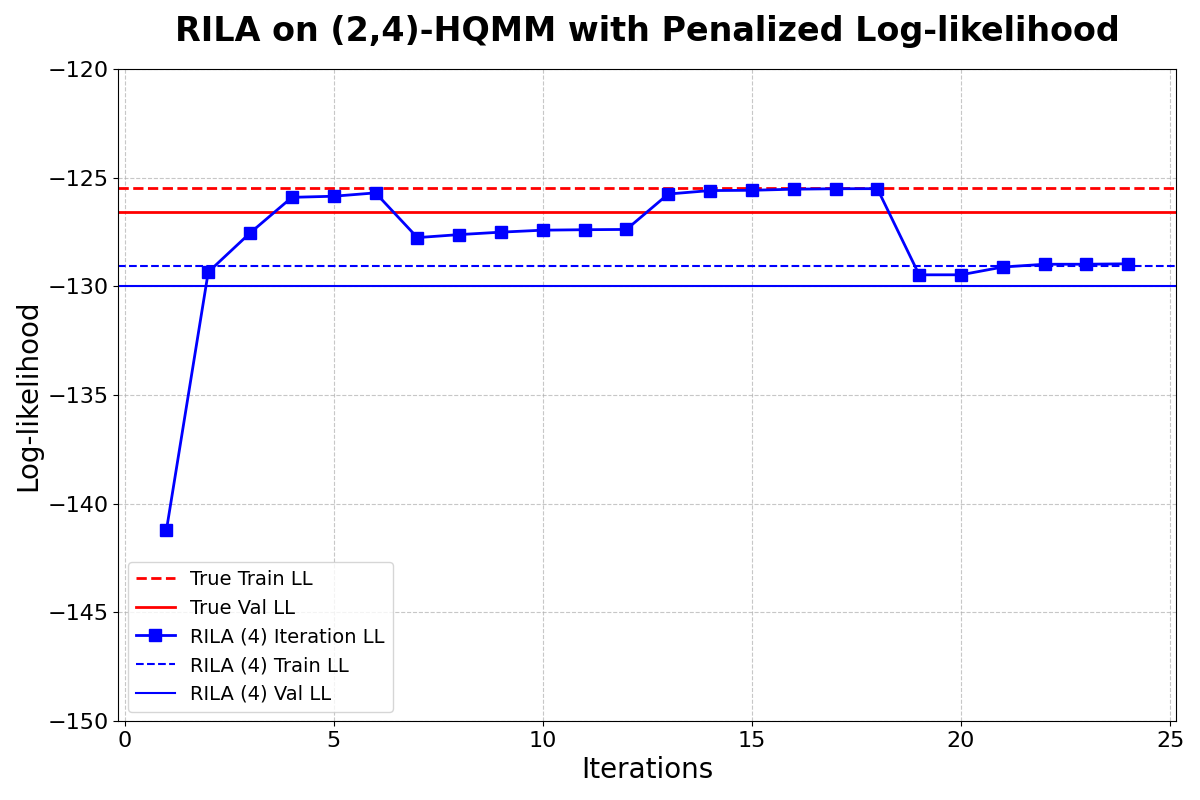}
	\caption{Line plot showing the 4-batch RILA performance in (2,4)-HQMM learning under the penalized log-likelihood. The RILA 4-batch iteration log-likelihoods are plotted as a blue line with square markers. The final RILA training (blue dashed) and validation (blue solid) log-likelihoods are shown as horizontal lines, while the true training (red dashed) and validation (red solid) log-likelihoods are overlaid for comparison.
	}	\label{fig:HQMM_2_4_iterations_P}
\end{figure}

\begin{figure}[t!]
	\centering
	\includegraphics[width=0.73\textwidth]{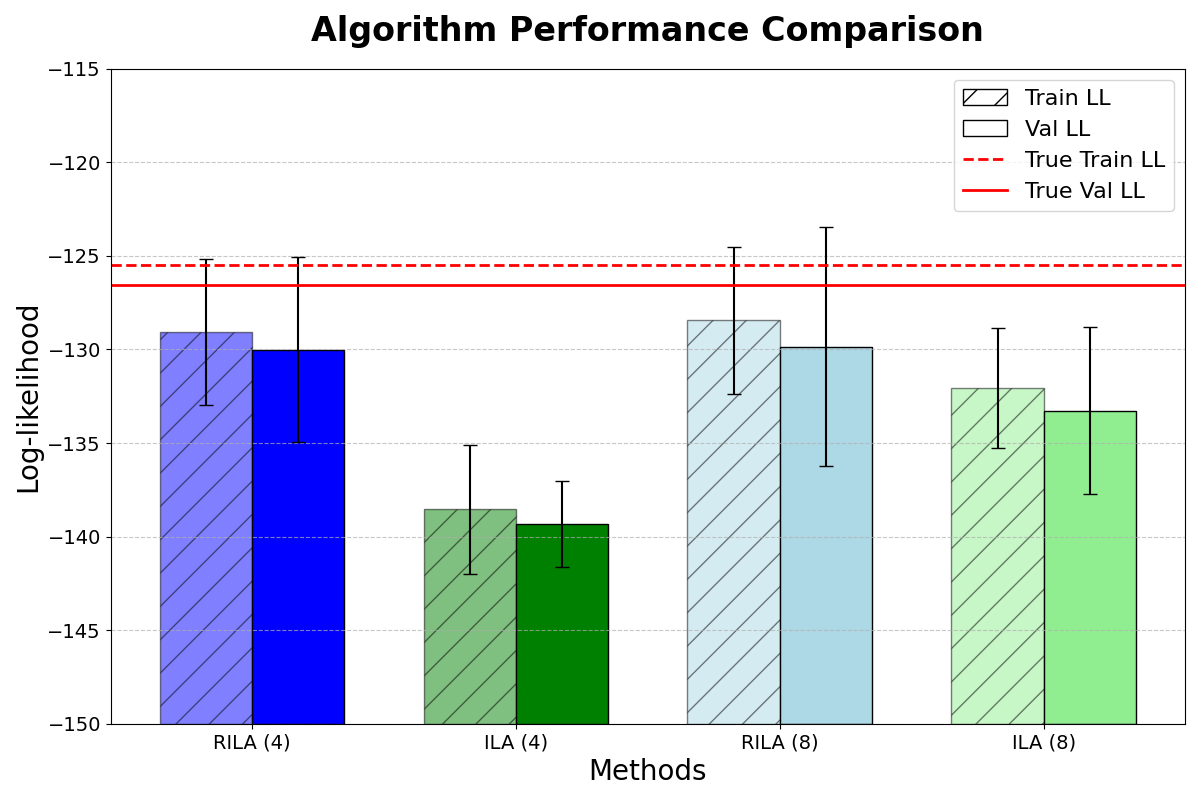}
	\caption{Grouped bar plot comparing the performance of RILA (4 and 8 batches) and ILA (4 and 8 batches) on a dataset generated from the (2,4)-HQMM under the penalized log-likelihood. Each bar represents the mean log-likelihood, with training (dashed-pattern, lighter bars) and validation (solid, darker bars) values shown side by side for each method. Error bars indicate standard deviations. True training (red dashed) and validation (red solid) log-likelihoods are overlaid as horizontal reference lines. }	
	\label{fig:Algorithm_Performance_Comparison_P}
\end{figure}

Figure \ref{fig:HQMM_2_4_iterations_P} illustrates the convergence behavior of the 4-batch RILA algorithm applied to the (2,4)-HQMM under the penalized log-likelihood formulation \eqref{eqn:LL_P}. The iteration-wise log-likelihoods exhibit a rapid initial improvement followed by stabilization after only a few iterations, indicating efficient convergence. The horizontal blue dashed and solid lines represent the final training and validation log-likelihoods obtained by RILA using the \texttt{patternsearch} optimizer, while the red dashed and solid lines indicate the corresponding true values. The results demonstrate that \texttt{patternsearch}-based RILA effectively maximizes the penalized objective and achieves log-likelihoods close to the true levels, confirming both stability and robustness of the optimization under regularization. We note that the ILA algorithm and code provided in \cite{srinivasan2018learning} are not compatible with the penalized log-likelihood, so its training results using the regular log-likelihood are not reported. Figure~\ref{fig:Algorithm_Performance_Comparison_P} presents a grouped bar plot comparing the performance of RILA (with 4 and 8 batches) and ILA (with 4 and 8 batches) using the penalized log-likelihood. The plot shows the mean log-likelihood for training and validation for each method, with error bars indicating standard deviations. The results indicate that while increasing the batch size to 8 enhances performance for ILA, RILA achieves the highest mean log-likelihood with 4 batches already.

\begin{figure}[t!]
	\centering
	\includegraphics[width=0.73\textwidth]{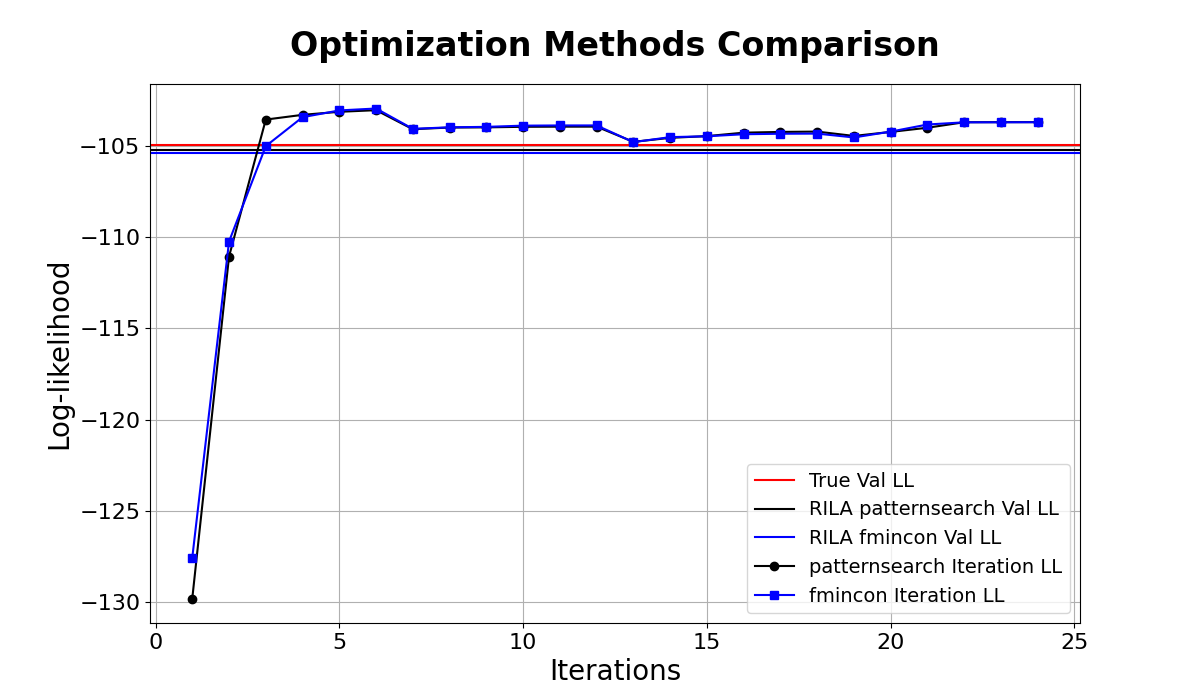}
	\caption{Comparison of HQMM learning performance using \texttt{patternsearch} and \texttt{fmincon} optimization methods, trained on a corrupted dataset generated from the (2,4)-HQMM. The plot shows the true validation log-likelihood (red solid), alongside RILA learned validation values for \texttt{patternsearch} (black solid) and \texttt{fmincon} (blue solid). Iteration log-likelihoods are depicted with black circles (\texttt{patternsearch}) and blue squares (\texttt{fmincon}) across four batches.}
	\label{fig:pattersearch_fmincon_C}
\end{figure}

\begin{figure}[t!]
	\centering
	\includegraphics[width=0.73\textwidth]{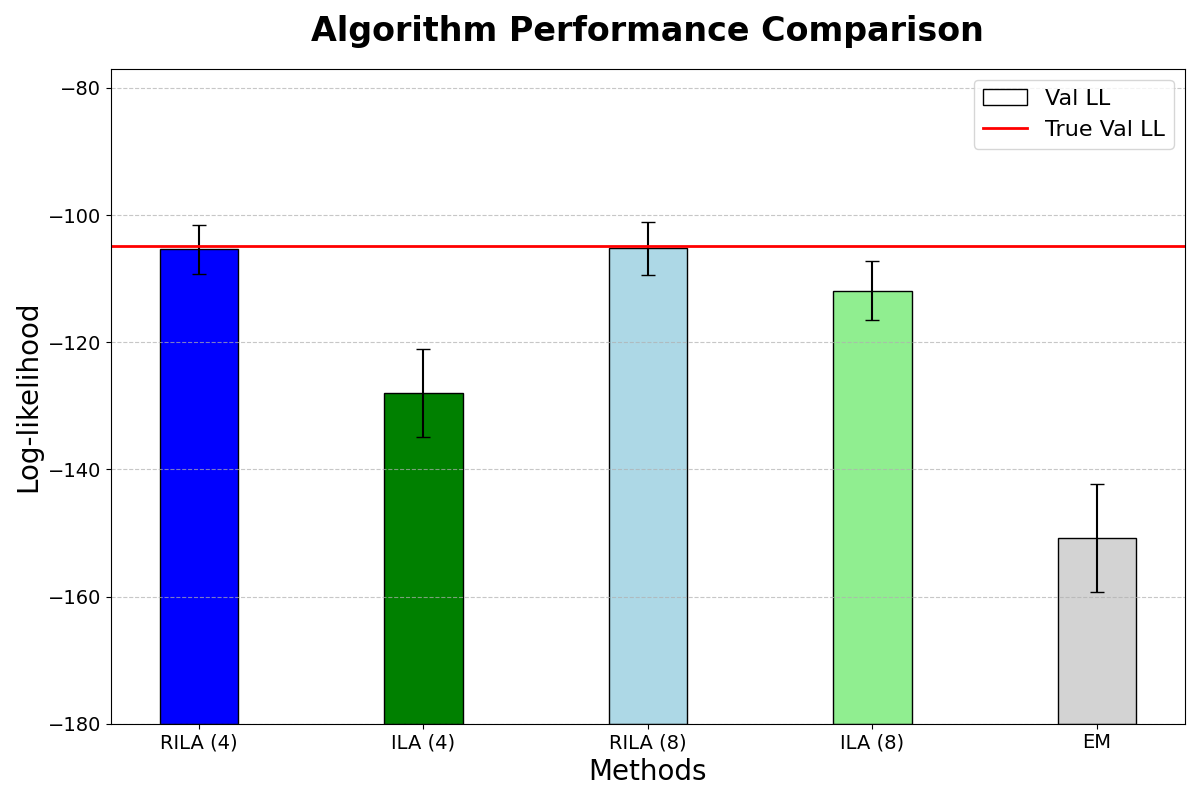}
	\caption{Grouped bar plot comparing the performance of RILA (4 and 8 batches), ILA (4 and 8 batches), and EM on a corrupted dataset generated from the (2,4)-HQMM. Each bar represents the mean log-likelihood, with training (dashed-pattern, lighter bars) and validation (solid, darker bars) values shown side by side for each method. Error bars indicate standard deviations. The true validation log-likelihood (red solid line) is shown as a horizontal reference.}	\label{fig:Algorithm_Performance_Comparison_C}
\end{figure}

Figure~\ref{fig:pattersearch_fmincon_C} presents a comparison of the learning performance of a HQMM using the \texttt{patternsearch} and \texttt{fmincon} optimization methods, trained on a corrupted dataset where 10 out of 30 sequences of the orginial training dataset are replaced by a value of 4. The plot displays the validation log-likelihood  alongside the RILA-learned validation log-likelihoods for \texttt{patternsearch} and \texttt{fmincon}, which are true values as the validation dataset is clean. Both methods show a rapid initial increase in log-likelihood, stabilizing around -105 after approximately 5 iterations, closely approaching the true validation log-likelihood. 
Figure~\ref{fig:Algorithm_Performance_Comparison_C} presents a grouped bar plot comparing the performance of RILA (with 4 and 8 batches), ILA (with 4 and 8 batches), and EM on the corrupted dataset. Given the corrupted training dataset, only the log-likelihoods of the validation datasets are reported, along with the true validation log-likelihood as a horizontal reference. The plot shows the mean validation log-likelihood for each method, with error bars indicating standard deviations. RILA is very close to the true validation log-likelihood, while the performance of other methods, such as EM, degrades significantly, with log-likelihoods dropping below -140, highlighting RILA's robustness on the corrupted data.

\begin{figure}[t!]
	\centering
	\includegraphics[width=0.73\textwidth]{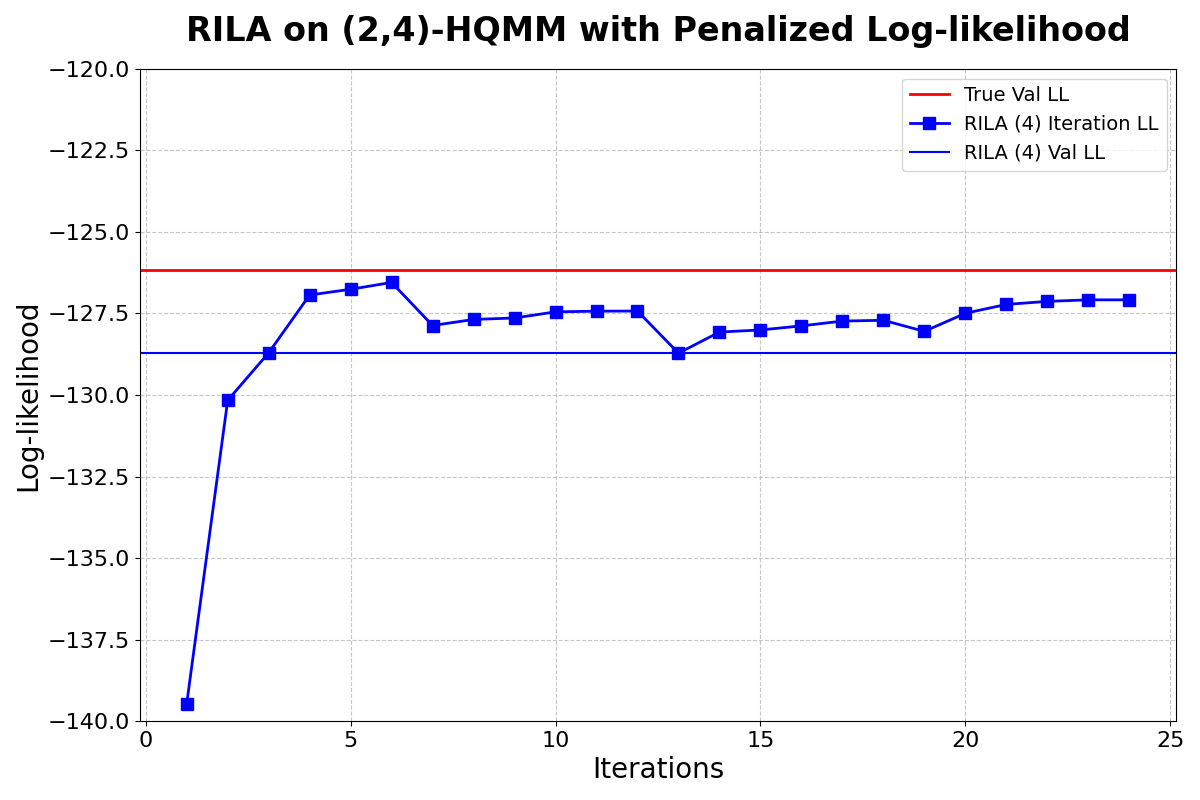}
	\caption{Line plot showing performance of RILA with penalized log-likelihood, on a corrupted dataset generated from the (2,4)-HQMM.
		The RILA 4-batch iteration log-likelihoods are plotted as a blue line with square markers. The final RILA validation (blue solid) and the true validation (red solid) log-likelihoods are shown as horizontal lines.
	}	\label{fig:HQMM_2_4_iterations_P_C}
\end{figure}

\begin{figure}[t!]
	\centering
	\includegraphics[width=0.73\textwidth]{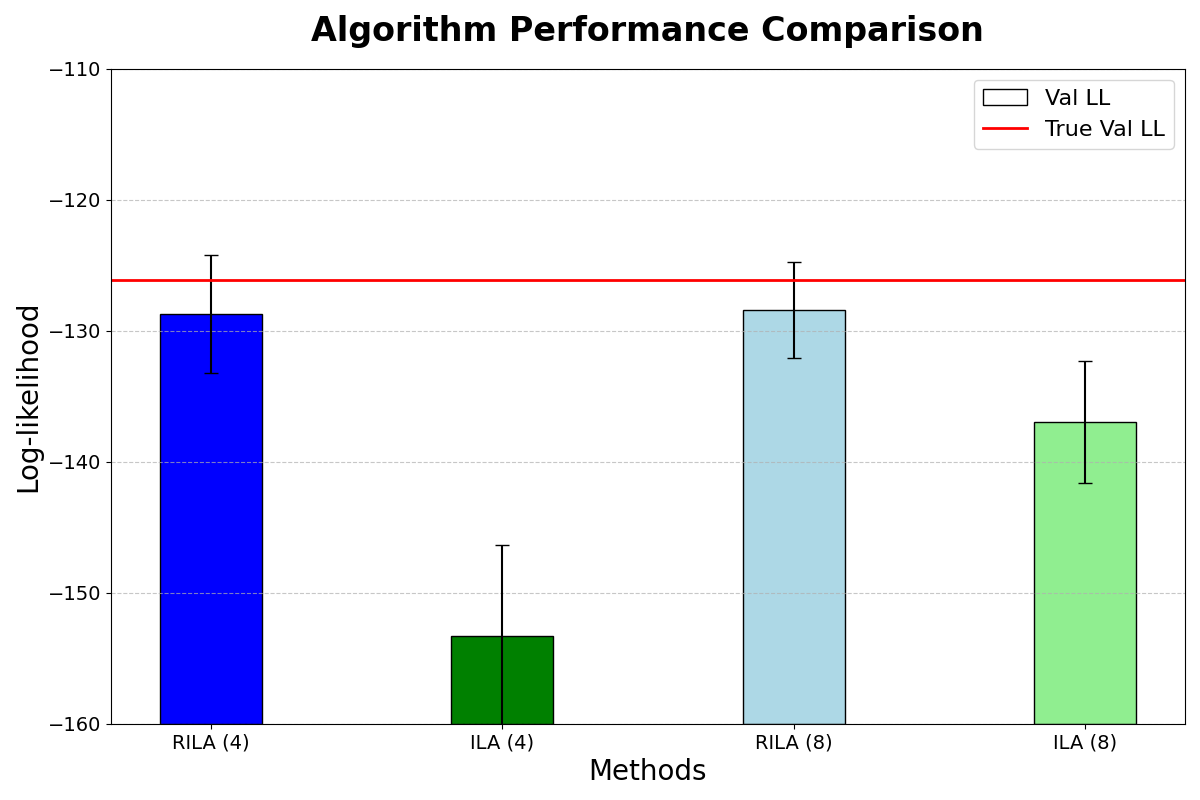}
	\caption{Grouped bar plot comparing the performance of RILA (4 and 8 batches) and ILA (4 and 8 batches) on a corrupted dataset generated from the (2,4)-HQMM under the penalized log-likelihood. Each bar represents the mean log-likelihood for each method. Error bars indicate standard deviations. The true validation log-likelihood (red solid line) is shown as a horizontal reference. }	
	\label{fig:Algorithm_Performance_Comparison_P_C}
\end{figure}

Figure~\ref{fig:HQMM_2_4_iterations_P_C} presents a line plot illustrating the performance of RILA with penalized log-likelihood on the corrupted dataset, using the \texttt{patternsearch} optimization method with 4 batches. Given the corrupted training dataset, only the validation log-likelihoods are reported. The RILA 4-batch iteration log-likelihoods start from around -140 and rise to stabilize near -127.5 after 5 iterations, indicating effective convergence despite the dataset corruption. Figure~\ref{fig:Algorithm_Performance_Comparison_P_C} presents a grouped bar plot comparing the performance of RILA (with 4 and 8 batches) and ILA (with 4 and 8 batches) on the corrupted dataset, utilizing the penalized log-likelihood. The plot displays the mean validation log-likelihood for each method, with error bars indicating standard deviations. RILA with 4 and 8 batches achieves the highest mean log-likelihood, closest to the true value at around -130, while ILA with 4 batches performs the poorest, dropping below -150, and increasing batch sizes to 8 improves performance for ILA but still falls far short of the true value.

\subsection{HQMM Example  in \cite{srinivasan2018learning}}
\label{sec:HQMM2}

\cite{srinivasan2018learning} gave the motivation of the 2-state HQMM by drawing inspiration from a physical system, specifically electron spin, which represents quantized angular momentum with two possible states: `up' or `down' along any chosen measurement axis, without intermediate values. Unlike classical angular momentum, which can be described by a vector with defined components across three spatial dimensions, electron spin lacks a well-defined 3D vector and is instead characterized as `up' or `down' relative to a selected measurement basis. Selecting the $z$-axis as a reference, the electron's spin state in the $\{ +z, - z\}$ basis can be expressed as $\begin{bmatrix} 1 & 0 \end{bmatrix}^T$ for $|+z\rangle$ and $\begin{bmatrix} 0 & 1 \end{bmatrix}^T$ for $|-z\rangle$. As a two-state quantum system, electron spin can exist in superpositions of these orthogonal `up' and `down' states, parameterized by $(\theta,\phi)$ and written as $$|\psi\rangle = \cos\left(\frac{\theta}2\right)|+z\rangle + e^{i\phi}\sin\left(\frac{\theta}2\right)|-z\rangle,$$ where $0 \leq \theta \leq \pi$ and $0 \leq \phi \leq 2\pi$. The Bloch sphere, a unit-radius sphere, serves as a valuable visualization tool for qubits, mapping any two-state system to a point on its surface using $(\theta, \phi)$ as polar and azimuthal angles. Alternative bases such as $\{ +x, - x\}$ or $\{ +y, - y\}$ can also be represented in the original basis:
\begin{alignat*}{3}
	|+x\rangle &= \frac1{\sqrt{2}}|+z\rangle + \frac1{\sqrt{2}}|-z\rangle, \hspace{1cm}&&\left(\theta=\frac{\pi}2, \phi = 0\right),\\
	|-x\rangle &= \frac1{\sqrt{2}}|+z\rangle - \frac1{\sqrt{2}}|-z\rangle, \hspace{1cm}&&\left(\theta=\frac{\pi}2, \phi = \pi\right),\\
	|+y\rangle &= \frac1{\sqrt{2}}|+z\rangle + \frac{i}{\sqrt{2}}|-z\rangle, \hspace{1cm}&&\left(\theta=\frac{\pi}2, \phi = \frac{\pi}2\right),\\
	|-y\rangle &= \frac1{\sqrt{2}}|+z\rangle - \frac{i}{\sqrt{2}}|-z\rangle, \hspace{1cm}&&\left(\theta=\frac{\pi}2, \phi = \frac{3\pi}2\right).
\end{alignat*}
This process draws inspiration from the Stern-Gerlach experiment in quantum mechanics. 
We start with an electron spin represented in the $\{+ z, - z\}$ basis. At each time step, we randomly and uniformly select one of the $x$, $y$, or $z$ directions and apply an inhomogeneous magnetic field along that axis, collapsing the electron spin to either `up' or `down' along that axis and deflecting the electron accordingly. The measurement outcomes are encoded as follows: $1$: $+ z$, $2$: $- z$, $3$: $+ x$, $4$: $- x$, $5$: $+ y$, $6$: $- y$. This indicates the measured axis and the spin state (`up' or `down') at each step. 
Transitions between $1 \leftrightarrow 2$, $3 \leftrightarrow 4$, and $5 \leftrightarrow 6$ are prohibited, as our process does not include spin-flip operations. 

\cite{srinivasan2018learning} added two new Kraus operators with complex entries to the (2,4)-HQMM, and then renormalize:
\begin{alignat*}{3}
	\hat{K}_{1,1} &= \begin{pmatrix} \frac1{\sqrt{3}} & 0 \vspace{0.08cm}\\ 0 & 0\end{pmatrix}, \hspace{1cm}
	&&\hat{K}_{2,1} = \begin{pmatrix} 0 & 0 \vspace{0.08cm}\\ 0 & \frac1{\sqrt{3}}\end{pmatrix}, \\
	\hat{K}_{3,1} &= \begin{pmatrix} \frac1{2\sqrt{3}} & \frac1{2\sqrt{3}} \vspace{0.08cm}\\ \frac1{2\sqrt{3}} & \frac1{2\sqrt{3}}\end{pmatrix},\hspace{1cm}
	&&\hat{K}_{4,1} = \begin{pmatrix} \frac1{2\sqrt{3}} & -\frac1{2\sqrt{3}}\vspace{0.05cm} \vspace{0.08cm}\\ -\frac1{2\sqrt{3}} & \frac1{2\sqrt{3}}\end{pmatrix},\\
	\hat{K}_{5,1} &= \begin{pmatrix} \frac1{2\sqrt{3}} & -\frac{i}{2\sqrt{3}} \vspace{0.08cm}\\ \frac{i}{2\sqrt{3}} & \frac1{2\sqrt{3}}\end{pmatrix}, \hspace{1cm}
	&&\hat{K}_{6,1} = \begin{pmatrix} \frac1{2\sqrt{3}} & \frac{i}{2\sqrt{3}} \vspace{0.08cm}\\ -\frac{i}{2\sqrt{3}} & \frac1{2\sqrt{3}}\end{pmatrix}.
\end{alignat*}
Kraus operators $\hat{K}_{1,1}$ and $\hat{K}_{2,1}$ maintain the spin along the $z$-axis, Kraus operators $\hat{K}_{3,1}$ and $\hat{K}_{4,1}$ adjust the spin to align with the $x$-axis, and Kraus operators $\hat{K}_{5,1}$ and $\hat{K}_{6,1}$ reorient the spin to the $y$-axis. 

\begin{figure}[t!]
	\centering
	\includegraphics[width=0.73\textwidth]{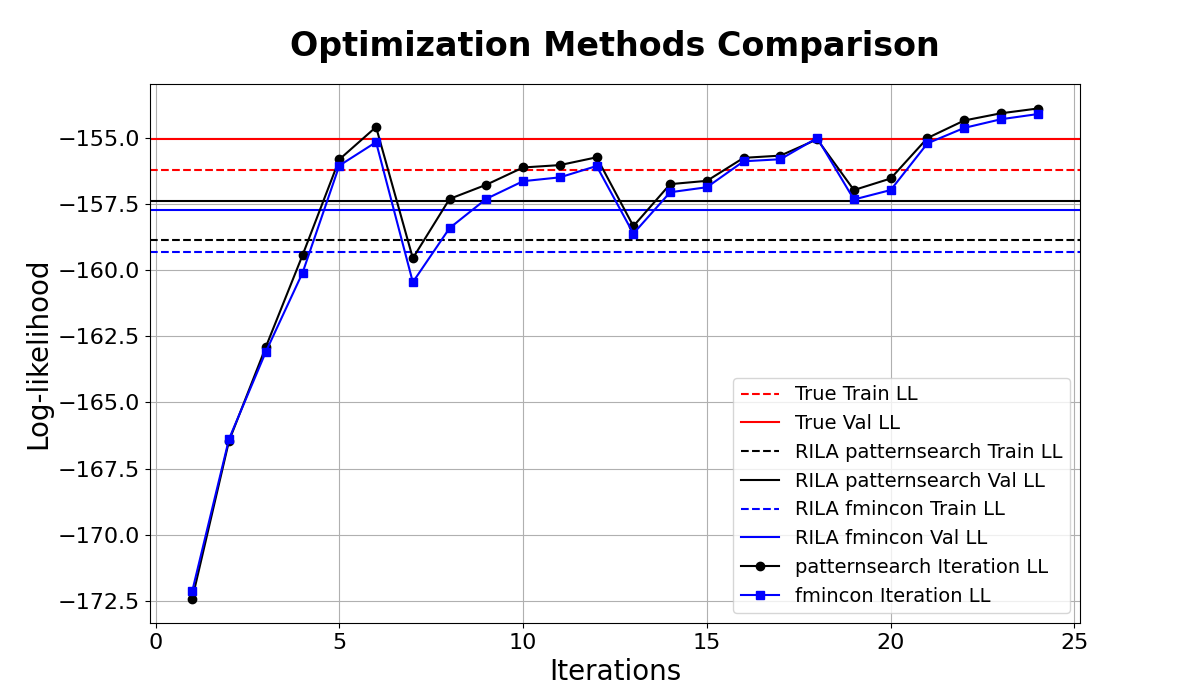}
	\caption{Comparison of HQMM learning performance using \texttt{patternsearch} and \texttt{fmincon} optimization methods on a dataset generated from the (2,6)-HQMM. 
		The plot shows true training (red dashed) and validation (red solid) log-likelihoods, alongside RILA learned values for \texttt{patternsearch} (black dashed for training, solid for validation) and \texttt{fmincon} (blue dashed for training, solid for validation). Iteration log-likelihoods are depicted with black circles (\texttt{patternsearch}) and blue squares (\texttt{fmincon}) across four batches.}
	\label{fig:HQMM_2_6_pattersearch_fmincon}
\end{figure}

\begin{figure}[t!]
	\centering
	\includegraphics[width=0.73\textwidth]{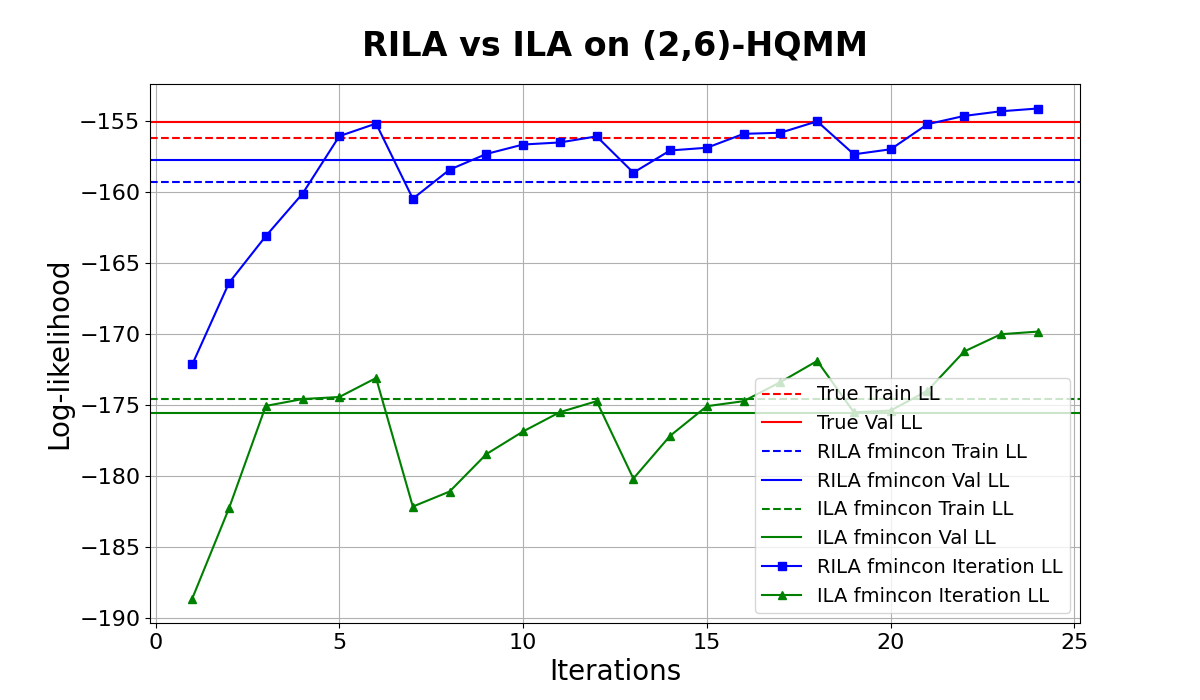}
	\caption{Comparison of HQMM learning performance using RILA and ILA across four batches on a dataset generated from the (2,6)-HQMM. The plot shows true training (red dashed) and validation (red solid) log-likelihoods, alongside RILA  learned values (blue dashed for training, solid for validation) and ILA  learned values (green dashed for training, solid for validation). Iteration log-likelihoods are depicted with blue squares (RILA) and green triangles (ILA) across four batches.}	\label{fig:HQMM_2_6_iterations2}
\end{figure}

\begin{figure}[t!]
	\centering
	\includegraphics[width=0.73\textwidth]{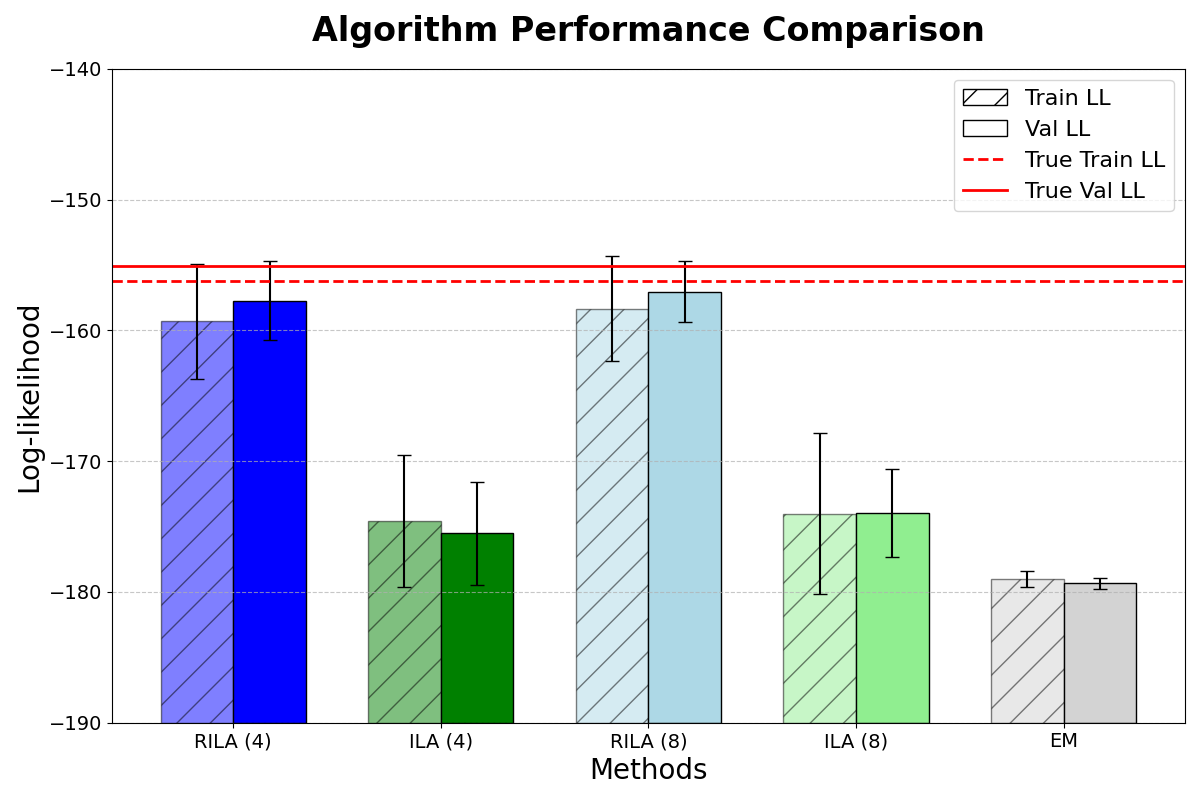}
	\caption{Grouped bar plot comparing the performance of RILA (4 and 8 batches), ILA (4 and 8 batches), and EM on a dataset generated from the (2,6)-HQMM. Each bar represents the mean log-likelihood, with training (dashed-pattern, lighter bars) and validation (solid, darker bars) values shown side by side for each method. Error bars indicate standard deviations. True training (red dashed) and validation (red solid) log-likelihoods are overlaid as horizontal reference lines. }	\label{fig:Algorithm_Performance_Comparison2}
\end{figure}

Figure~\ref{fig:HQMM_2_6_pattersearch_fmincon} presents a comparison of the learning performance of RILA using the \texttt{patternsearch} and \texttt{fmincon} optimization methods on a dataset generated from the (2,6)-HQMM. The plot shows the true training log-likelihood and validation log-likelihood, both stabilizing around -155, alongside the RILA-learned log-likelihoods for \texttt{patternsearch} and \texttt{fmincon}. Iteration log-likelihoods exhibit a rapid initial rise from -172.5, with both optimization methods delivering nearly identical performance. Figure~\ref{fig:HQMM_2_6_iterations2} presents a comparison of the learning performance of RILA and ILA across four batches on this dataset. The plot displays the true training log-likelihood and validation log-likelihood, both stabilizing around -155, alongside the RILA-learned and ILA-learned log-likelihoods. Iteration log-likelihoods show a rapid initial increase from approximately -190 for ILA and -175 for RILA, with RILA converging closer to the true validation log-likelihood around -157 after 5 iterations, while ILA stabilizes slightly lower at around -175, indicating RILA's superior performance in this scenario. Figure~\ref{fig:Algorithm_Performance_Comparison2} presents a grouped bar plot comparing the performance of RILA (with 4 and 8 batches), ILA (with 4 and 8 batches), and EM on this dataset. The plot shows the mean log-likelihood, with training and validation values for each method. The true training and validation log-likelihoods, both around -155, serve as horizontal references. 
RILA with 8 batches achieves the highest mean validation log-likelihood, approaching -155, and performs slightly better than RILA with 4 batches. Increasing the number of training batches leads to only a very small improvement for ILA. As expected, EM performs poorly, with validation log-likelihoods dropping below -170. Overall, RILA significantly outperforms ILA in this (2,6)-HQMM learning task, showing a larger performance gap than observed for the (2,4)-HQMM in Section \ref{sec:HQMM1}. In contrast, the advantage of ILA over EM in this task is much smaller than in the (2,4)-HQMM case.

\begin{figure}[t!]
	\centering
	\includegraphics[width=0.73\textwidth]{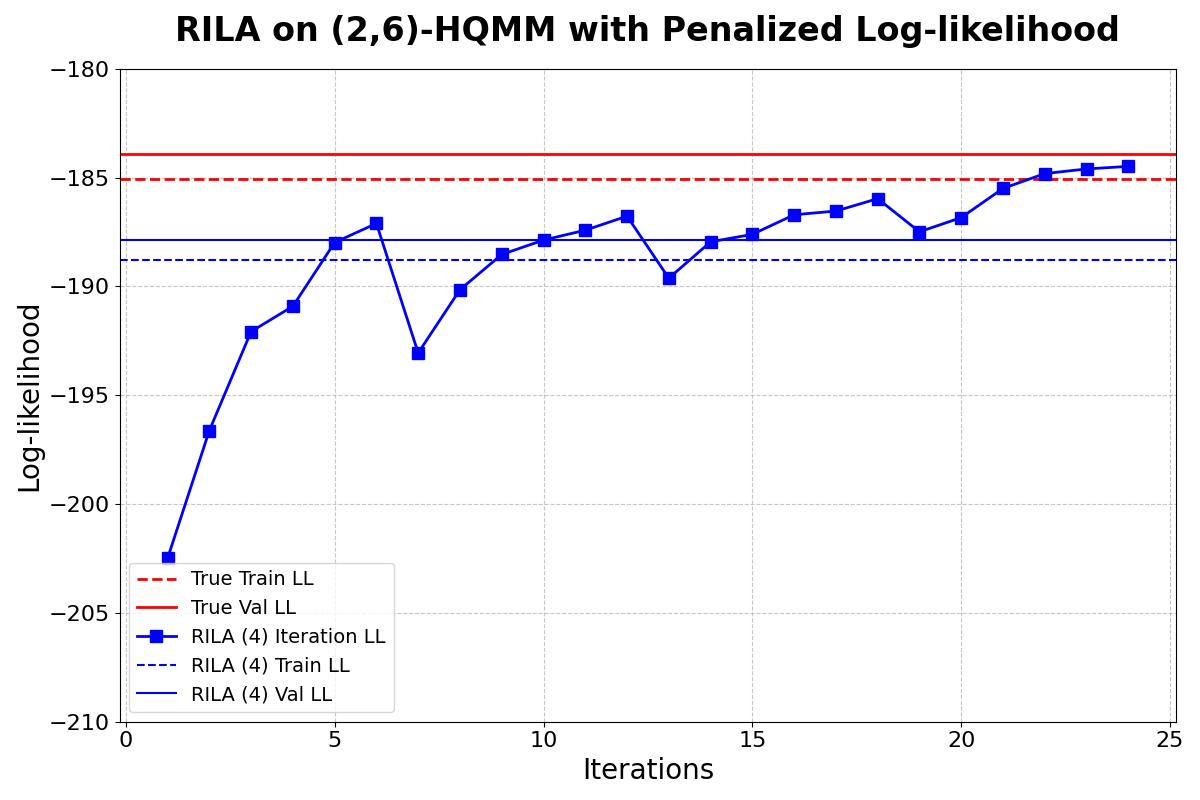}
	\caption{Line plot showing the 4-batch RILA performance in the (2,6)-HQMM learning with penalized log-likelihood. The RILA 4-batch iteration log-likelihoods are plotted as a blue line with square markers. The final RILA training (blue dashed) and validation (blue solid) log-likelihoods are shown as horizontal lines, while the true training (red dashed) and validation (red solid) log-likelihoods are overlaid for comparison.
	}	\label{fig:HQMM_2_6_iterations_P}
\end{figure}

\begin{figure}[t!]
	\centering
	\includegraphics[width=0.73\textwidth]{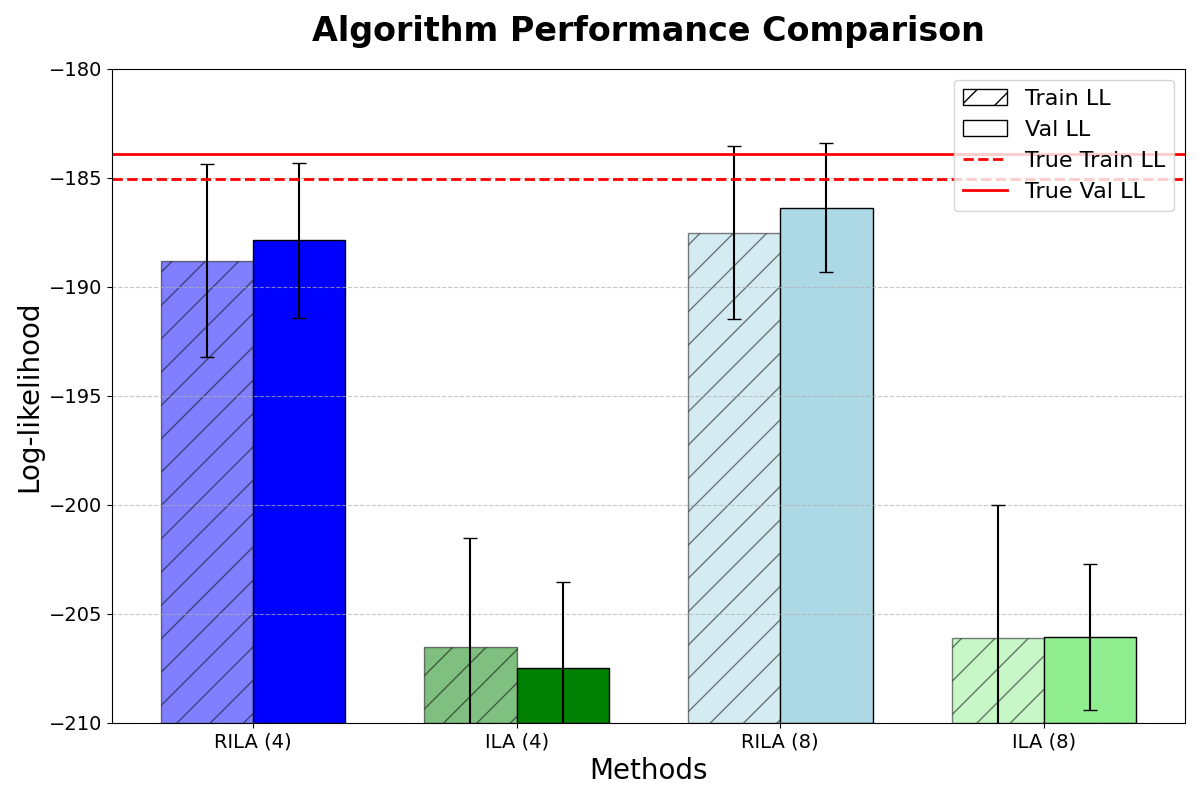}
	\caption{Grouped bar plot comparing the performance of RILA (4 and 8 batches) and ILA (4 and 8 batches) on a dataset generated from the (2,6)-HQMM under the penalized log-likelihood. Each bar represents the mean log-likelihood, with training (dashed-pattern, lighter bars) and validation (solid, darker bars) values shown side by side for each method. Error bars indicate standard deviations. True training (red dashed) and validation (red solid) log-likelihoods are overlaid as horizontal reference lines. }	
	\label{fig:HQMM_2_6_Algorithm_Performance_Comparison_P}
\end{figure}

\begin{figure}[t!]
	\centering
	\includegraphics[width=0.73\textwidth]{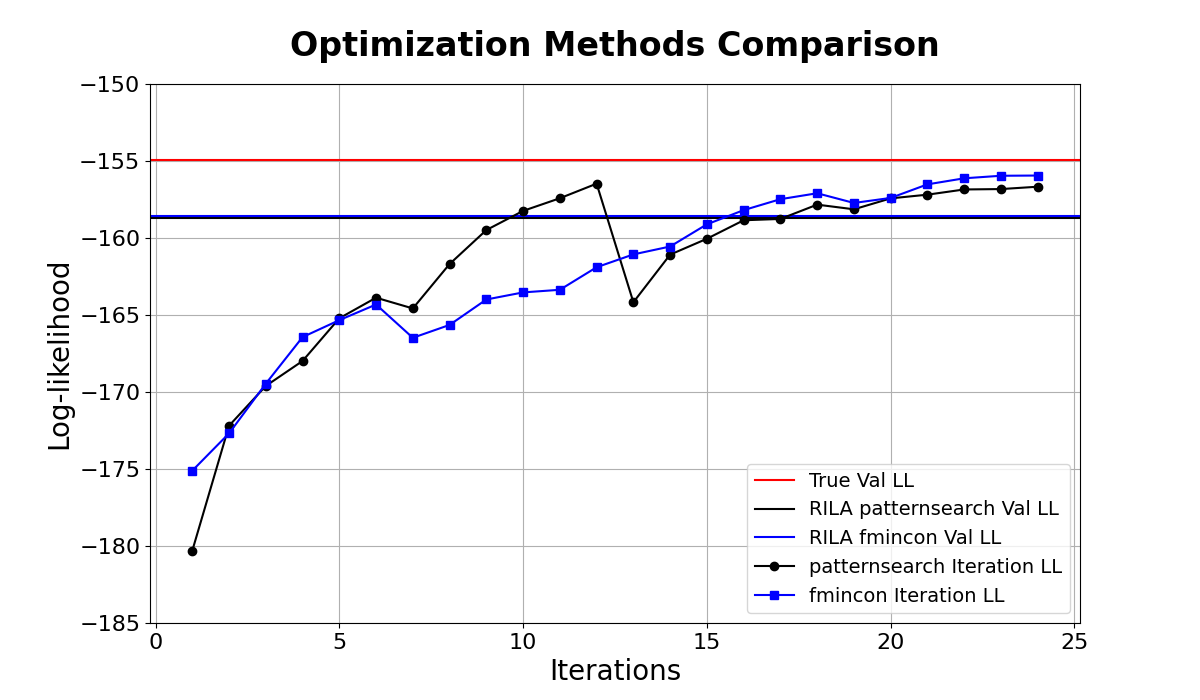}
	\caption{Comparison of HQMM learning performance using \texttt{patternsearch} and \texttt{fmincon} optimization methods, trained on a corrupted dataset generated from the (2,6)-HQMM.
		The plot shows the true validation log-likelihood (red solid), alongside RILA learned validation values for \texttt{patternsearch} (black solid) and \texttt{fmincon} (blue solid). Iteration log-likelihoods are depicted with black circles (\texttt{patternsearch}) and blue squares (\texttt{fmincon}) across four batches.}
	\label{fig:HQMM_2_6_pattersearch_fmincon_C}
\end{figure}

Figure \ref{fig:HQMM_2_6_iterations_P} illustrates the convergence behavior of the 4-batch RILA algorithm applied to the (2,6)-HQMM, under the penalized log-likelihood formulation. The iteration-wise log-likelihoods exhibit a rapid initial improvement  indicating efficient convergence. The horizontal blue dashed and solid lines represent the final training and validation log-likelihoods obtained by RILA using the \texttt{patternsearch} optimizer. The results demonstrate that \texttt{patternsearch}-based RILA effectively maximizes the penalized objective and achieves log-likelihoods close to the true levels, confirming both stability and robustness of the optimization under regularization.
Figure~\ref{fig:HQMM_2_6_Algorithm_Performance_Comparison_P} presents a grouped bar plot comparing the performance of RILA (with 4 and 8 batches) and ILA (with 4 and 8 batches) on this dataset, with penalized log-likelihood. The results show that increasing the batch size from 4 to 8 has no notable impact on either algorithm. Additionally, RILA's performance significantly outperforms ILA in terms of penalized log-likelihood.

\begin{figure}
	\centering
	\includegraphics[width=0.73\textwidth]{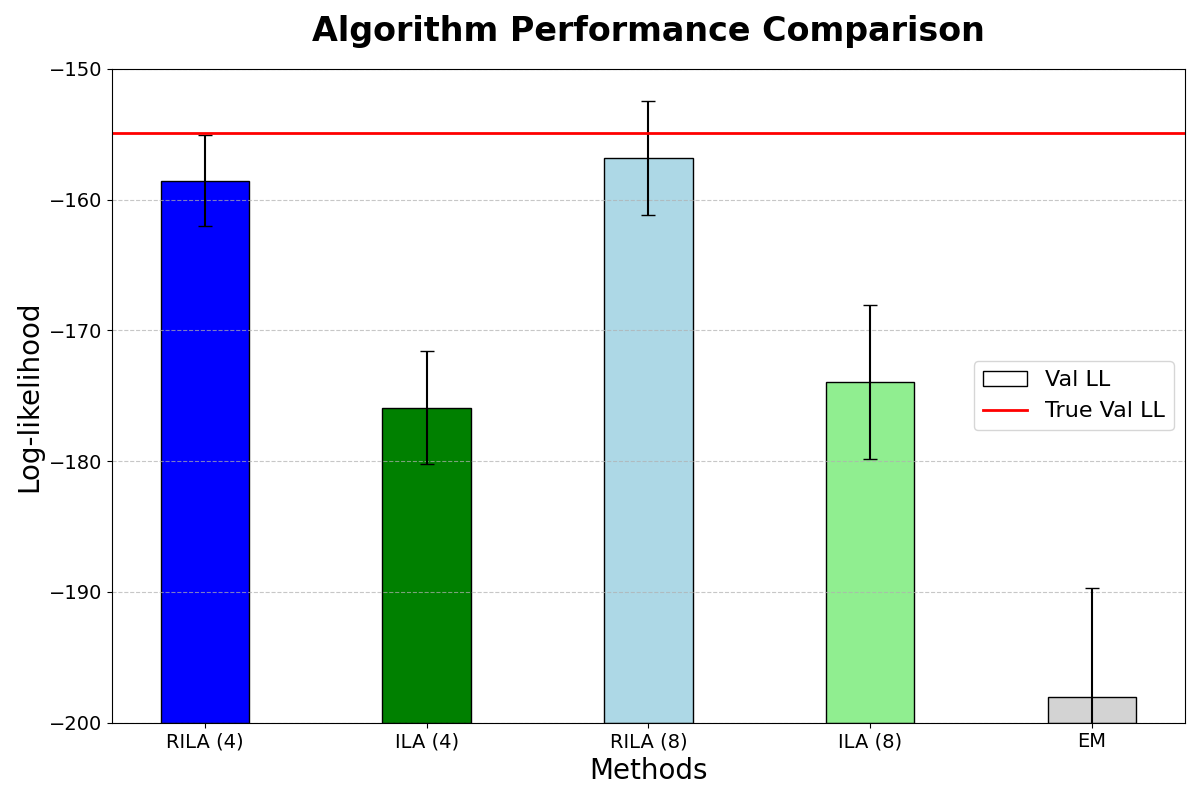}
	\caption{Grouped bar plot comparing the performance of RILA (4 and 8 batches), ILA (4 and 8 batches), and EM on a corrupted dataset generated from the (2,6)-HQMM. Each bar represents the mean log-likelihood, with training (dashed-pattern, lighter bars) and validation (solid, darker bars) values shown side by side for each method. The true validation log-likelihood (red solid line) is shown as a horizontal reference.}	\label{fig:HQMM_2_6_Algorithm_Performance_Comparison_C}
\end{figure}

\begin{figure}[t!]
	\centering
	\includegraphics[width=0.73\textwidth]{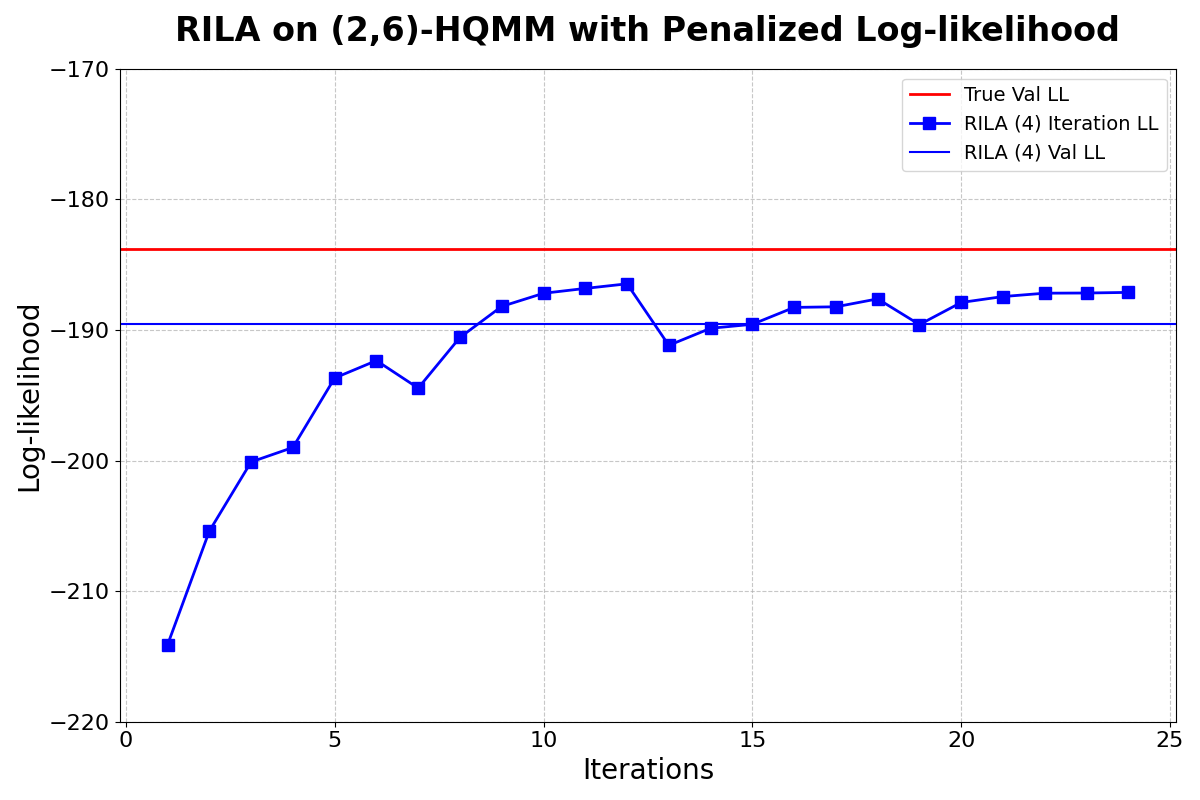}
	\caption{Line plot showing performance of RILA with penalized log-likelihood, on a corrupted dataset generate from the (2,6)-HQMM.
		The RILA 4-batch iteration log-likelihoods are plotted as a blue line with square markers. The final RILA validation (blue solid) and the true validation (red solid) log-likelihoods are shown as horizontal lines.
	}	\label{fig:HQMM_2_6_iterations_P_C}
\end{figure}

\begin{figure}[t!]
	\centering
	\includegraphics[width=0.73\textwidth]{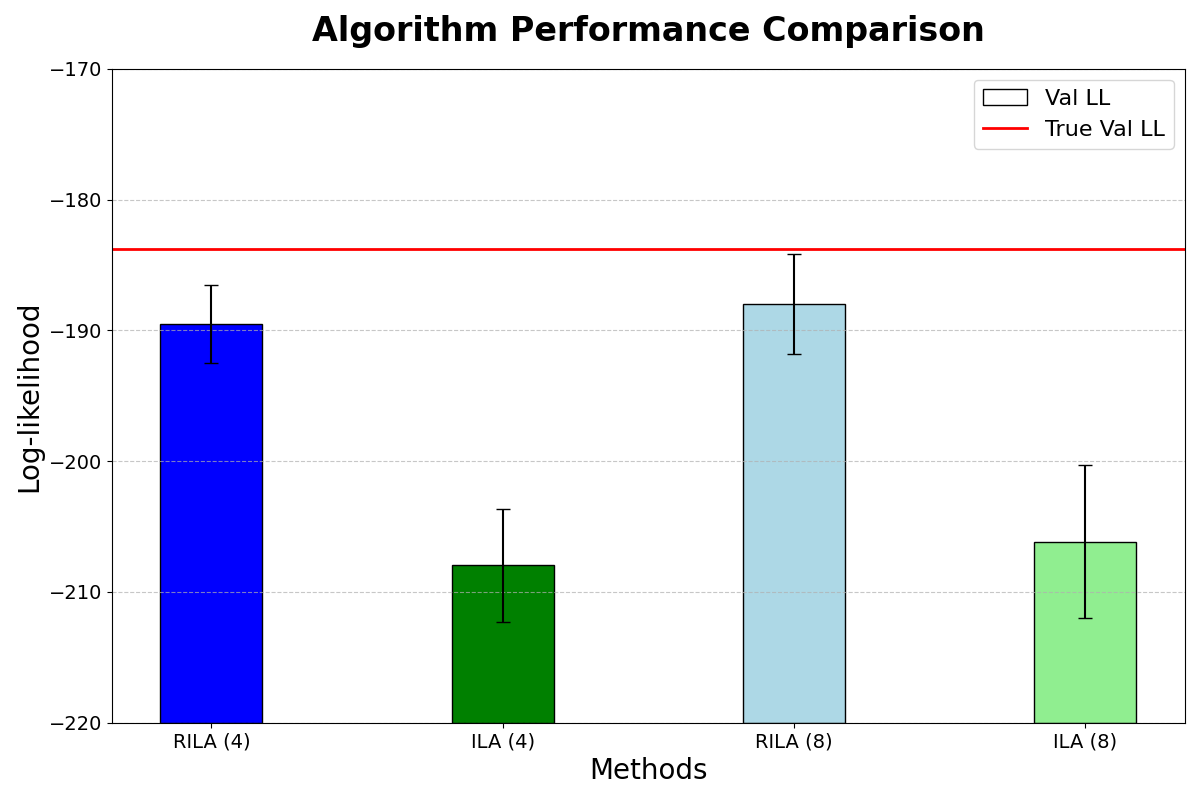}
	\caption{Grouped bar plot comparing the performance of RILA (4 and 8 batches) and ILA (4 and 8 batches) on a corrupted dataset generated from the (2,6)-HQMM under the penalized log-likelihood. Each bar represents the mean log-likelihood for each method. Error bars indicate standard deviations. The true validation log-likelihood (red solid line) is shown as a horizontal reference. }	
	\label{fig:HQMM_2_6_Algorithm_Performance_Comparison_P_C}
\end{figure}

Figure~\ref{fig:HQMM_2_6_pattersearch_fmincon_C} presents a comparison of the learning performance of a HQMM using the \texttt{patternsearch} and \texttt{fmincon} optimization methods, trained on a corrupted dataset where 10 out of 30 sequences are replaced by a value of 4. The plot displays the true validation log-likelihood  alongside the RILA-learned validation log-likelihoods utilizing both methods. Both methods show a compartively slower initial increase in log-likelihood, stabilizing around -160 after approximately 15 iterations, closely approaching the true validation log-likelihood. Figure~\ref{fig:HQMM_2_6_Algorithm_Performance_Comparison_C} presents a grouped bar plot comparing the performance of RILA (with 4 and 8 batches), ILA (with 4 and 8 batches), and EM on a corrupted dataset generated from the (2,4)-HQMM. 
RILA, with batch sizes of 4 and 8, performs closely to the true validation log-likelihood. In contrast, other methods, particularly EM, exhibit significant performance degradation, with log-likelihoods dropping to approximately -200.

Figure~\ref{fig:HQMM_2_6_iterations_P_C} presents a line plot illustrating the performance of RILA with penalized log-likelihood on a corrupted dataset generated from the (2,6)-HQMM, using the \texttt{patternsearch} optimization method with 4 batches. The RILA 4-batch iteration log-likelihoods are depicted as a blue line with square markers, starting from around -215 and rising to stabilize near -190 after 5 iterations, indicating effective convergence despite the dataset corruption. 
Figure~\ref{fig:HQMM_2_6_Algorithm_Performance_Comparison_P_C} presents a grouped bar plot comparing the performance of RILA (with 4 and 8 batches) and ILA (with 4 and 8 batches) on a corrupted dataset generated from the (2,6)-HQMM, using penalized log-likelihood. RILA with 8 batches achieves the highest mean log-likelihood, closest to the true value, while ILA with 4 batches performs the poorest, dropping below -250, and increasing batch sizes to 8 improves performance for ILA but still falls far short of the true value.


\subsection{HMM  Example}
\label{sec:HMM3}

We generated an (8,8)-HMM using the \texttt{hmmgenerate} function,  associated with MATLAB's Statistics and Machine Learning Toolbox to generate a sequence of observed emissions and their corresponding hidden states. The transition and emission matrices are generated by the \texttt{generateHMMParams} function provided in the code of \cite{srinivasan2018learning}. Specifically, the transition matrix is given by
\[
\mathbf{A} =
\begin{bmatrix}
	0.1039 & 0.1020 & 0.2531 & 0.2001 & 0.2169 & 0.1346 & 0.1579 & 0.0115 \\
	0.1410 & 0.1366 & 0.2584 & 0.1114 & 0.1641 & 0.0608 & 0.0404 & 0.1236 \\
	0.1097 & 0.0343 & 0.0246 & 0.1445 & 0.0615 & 0.0091 & 0.1621 & 0.1531 \\
	0.1794 & 0.0484 & 0.0113 & 0.0659 & 0.1731 & 0.3175 & 0.1925 & 0.1187 \\
	0.0535 & 0.1958 & 0.0490 & 0.1434 & 0.0226 & 0.0990 & 0.0282 & 0.2178 \\
	0.2298 & 0.2368 & 0.2536 & 0.1743 & 0.0982 & 0.1242 & 0.1139 & 0.1353 \\
	0.0072 & 0.0766 & 0.0284 & 0.0038 & 0.1992 & 0.2299 & 0.1910 & 0.2083 \\
	0.1755 & 0.1693 & 0.1216 & 0.1567 & 0.0644 & 0.0250 & 0.1139 & 0.0317 \\
\end{bmatrix}
\]
and the emisson matrix is given by
\[
\mathbf{C}  =
\begin{bmatrix}
	0.0327 & 0.1710 & 0.1649 & 0.2154 & 0.2030 & 0.1879 & 0.0064 & 0.0348 \\
	0.1894 & 0.1207 & 0.1545 & 0.1368 & 0.1393 & 0.1404 & 0.0084 & 0.0782 \\
	0.0933 & 0.1454 & 0.0285 & 0.0007 & 0.0035 & 0.0133 & 0.0091 & 0.1640 \\
	0.0388 & 0.0675 & 0.2360 & 0.1471 & 0.2077 & 0.1522 & 0.0791 & 0.2714 \\
	0.2176 & 0.0523 & 0.1118 & 0.0779 & 0.1544 & 0.1519 & 0.2762 & 0.1571 \\
	0.0816 & 0.1734 & 0.1438 & 0.1257 & 0.2229 & 0.1860 & 0.1730 & 0.0052 \\
	0.1762 & 0.0829 & 0.1015 & 0.2112 & 0.0385 & 0.1433 & 0.1775 & 0.2241 \\
	0.1704 & 0.1868 & 0.0589 & 0.0852 & 0.0306 & 0.0250 & 0.2704 & 0.0652 \\
\end{bmatrix}.
\]
The initial density matrix $\rho$ for a quantum state, derived from the initial state vector of HMM 
$$ v = \begin{bmatrix} 1 & 0 & 0 & 0 & 0 & 0 & 0 & 0 \end{bmatrix}^T,$$ 
can be expressed as the outer product of the state vector with its conjugate transpose. For a pure state, this is given by
$$\rho = |v\rangle\langle v| = \begin{bmatrix} 1 & 0 & 0 & 0 & 0 & 0 & 0 & 0 \\ 0 & 0 & 0 & 0 & 0 & 0 & 0 & 0 \\ 0 & 0 & 0 & 0 & 0 & 0 & 0 & 0 \\ 0 & 0 & 0 & 0 & 0 & 0 & 0 & 0 \\ 0 & 0 & 0 & 0 & 0 & 0 & 0 & 0 \\ 0 & 0 & 0 & 0 & 0 & 0 & 0 & 0 \\ 0 & 0 & 0 & 0 & 0 & 0 & 0 & 0 \\ 0 & 0 & 0 & 0 & 0 & 0 & 0 & 0 \end{bmatrix}.$$
This density matrix represents a pure state where the system is fully in the first basis state.

\begin{figure}[t!]
	\centering
	\includegraphics[width=0.73\textwidth]{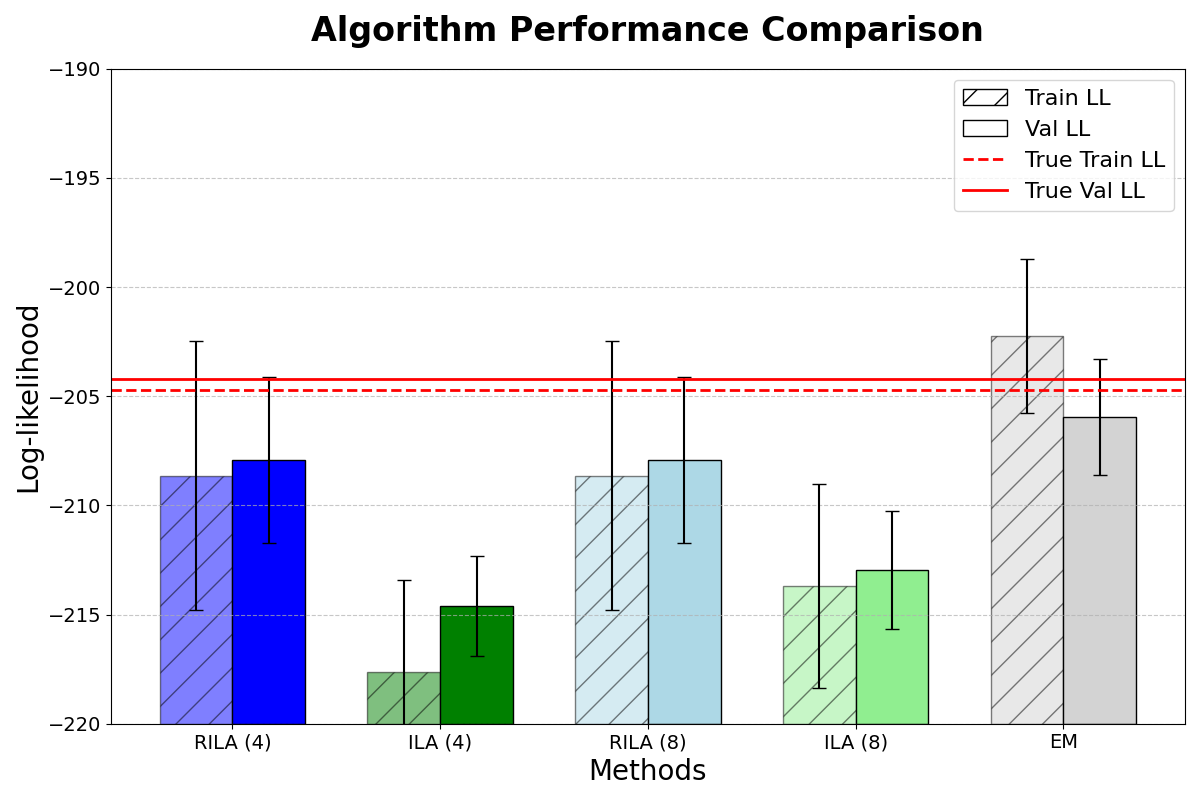}
	\caption{Grouped bar plot comparing the performance of RILA (4 and 8 batches), ILA (4 and 8 batches), and EM on a dataset generated from the (8,8)-HMM. Each bar represents the mean log-likelihood, with training (dashed-pattern, lighter bars) and validation (solid, darker bars) values shown side by side for each method. Error bars indicate standard deviations. True training (red dashed) and validation (red solid) log-likelihoods are overlaid as horizontal reference lines. }	\label{fig:HMM_8_8_Algorithm_Performance_Comparison}
\end{figure}

\begin{figure}[t!]
	\centering
	\includegraphics[width=0.73\textwidth]{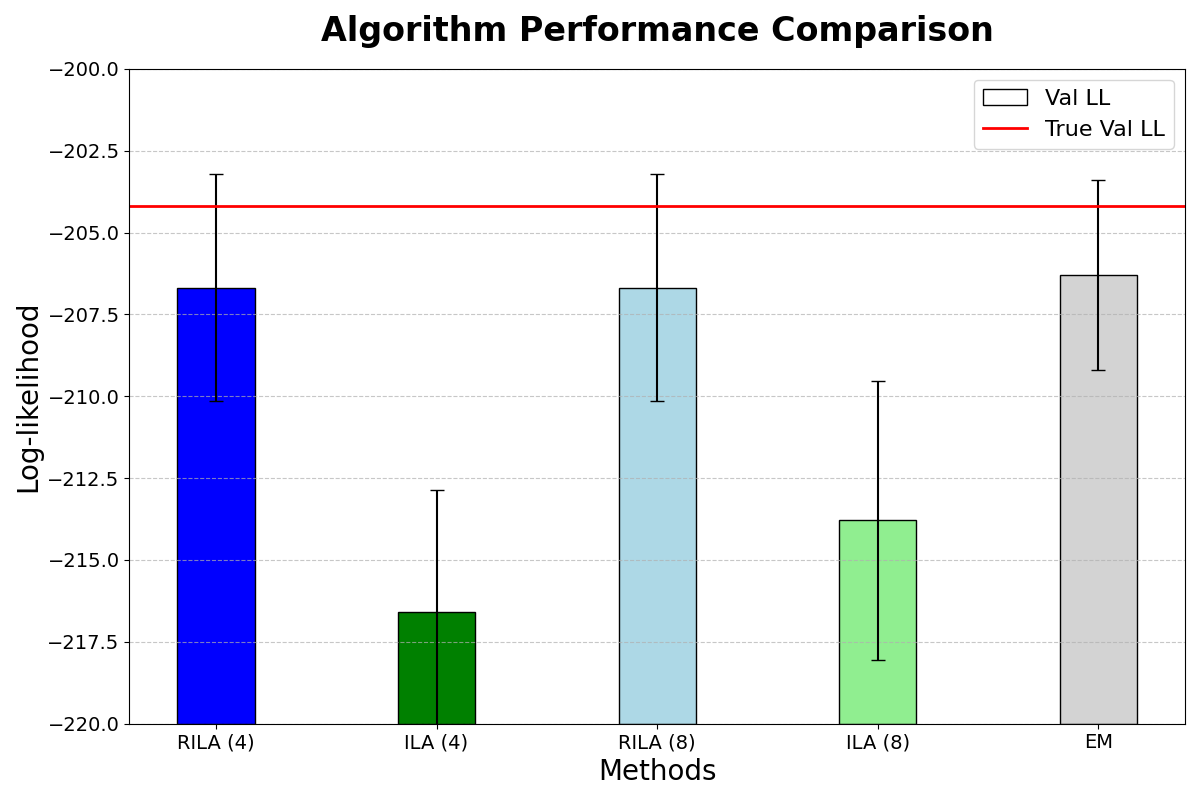}
	\caption{Grouped bar plot comparing the performance of RILA (4 and 8 batches), ILA (4 and 8 batches), and EM on a corrupted dataset generated from the (8,8)-HMM. Each bar represents the mean log-likelihood, with training (dashed-pattern, lighter bars) and validation (solid, darker bars) values shown side by side for each method. Error bars indicate standard deviations. The true validation log-likelihood (red solid line) is shown as a horizontal reference.}	\label{fig:HMM_8_8_Algorithm_Performance_Comparison_C}
\end{figure}

Figure~\ref{fig:HMM_8_8_Algorithm_Performance_Comparison} presents the performance results of RILA (with 4 and 8 batches), ILA (with 4 and 8 batches), and EM on a dataset generated from an (8,8)-HMM. None of the five methods yield highly accurate learning results. EM achieves the closest validation log-likelihood to the true value but at the cost of producing a much higher training log-likelihood than expected, which is unusual and not observed in previous HQMM learning experiments discussed in earlier sections. Increasing the number of training batches only slightly improves the performance of RILA and ILA, with RILA significantly outperforming ILA. Figure~\ref{fig:HMM_8_8_Algorithm_Performance_Comparison_C} compares the performance of these five methods on the same dataset, but with one-third of the sequences corrupted to take the value 4. Under this adversarial condition, RILA achieves a validation log-likelihood comparable to EM, while ILA's performance remains substantially lower.

Figures~\ref{fig:HMM_8_8_Algorithm_Performance_Comparison} and \ref{fig:HMM_8_8_Algorithm_Performance_Comparison_C} demonstrate that for datasets generated by classical models, the classical EM algorithm is preferable. Although RILA is designed for quantum-generalized HMMs, its performance is hindered by learning extraneous coherence. Furthermore, enforcing zero off-diagonal entries in the state density matrix is unnecessary, as this strategy increases algorithmic complexity. Thus, the best practice for classical model learning is to use the EM algorithm. However, when the dataset is corrupted, RILA's improved performance stems from its robust learning capability. In this case, RILA achieves comparable results on the corrupted dataset.

\section{Conclusion}
\label{Conclusion}

In this paper, we have introduced the AC-HQMM and the RILA, advancing the modeling and learning of quantum sequential systems in the presence of adversarial perturbations. The AC-HQMM framework effectively extends traditional HQMMs to manage a fraction of corrupted observation sequences while maintaining probabilistic consistency in the absence of corruption. Through comprehensive evaluations across clean and corrupted datasets using regular and penalized log-likelihood objectives, RILA demonstrates superior convergence stability, robustness to corruption, and preservation of physical validity compared to the ILA, particularly for quantum-generated data. The classical EM algorithm remains preferable for classical HMM datasets unless corrupted, where RILA's robust filtering and penalized optimization enable comparable performance. Applications of HQMMs have emerged in quantum process modeling, synthetic sequence generation, and even in classical time-series domains such as finance and biology~\citep{ma2023financial}. However, existing HQMM formulations and applications are limited to discrete-state representations. In practice, continuous-state HMMs—commonly known as state-space models—are widely used for complex dynamical systems (see, e.g.,~\cite{qiu2018multivariate, ning2023iterated}). Extending HQMMs to continuous-state formulations therefore remains an important open problem for future research. The code and data to reproduce our numerical results are publicly available at \url{https://github.com/patning/RILA}.

\appendix
\section{Proof of Proposition \ref{thm:diagonal_match}}
\label{proof:thm:diagonal_match}
\begin{proof}
Consider a general quantum system state
$$
\hat{\rho}_{t-1} = \begin{pmatrix} r & c \\ c^* & s \end{pmatrix}, \quad r + s = 1, \quad r, s \geq 0, \quad |c| \leq \sqrt{r s},
$$
where the diagonal corresponds to the classical belief state $x_{t-1} = (r, s)^T$.
\medskip

 \noindent\textbf{Step 1:} We first calculate the posterior distribution of the HMM. 
The classical belief state evolves via the transition matrix as
$$
A x_{t-1} = \begin{pmatrix} a r + (1-b) s \\ (1-a) r + b s \end{pmatrix}.
$$
From the emission matrix we know that, for $y\in\{1,2,3\}$,
$$P(y | j=0) = e_y\quad \text{and}\quad P(y | j=1) = f_y.$$  
The unnormalized posterior probability for observation $y$ in a classical HMM with hidden state $j$ at time $t$ is given by
\begin{equation}
	x_t(j \mid y) \propto P(y \mid j) \cdot (A x_{t-1})(j),
\end{equation}
which gives
\begin{equation}
	x_t(j \mid y)  \propto \left\{ \begin{array}{rcl}
		e_y [a r + (1-b) s] & \mbox{for }  j=0, \vspace{0.1cm}\\
		f_y [(1-a) r + b s] & \mbox{for }  j=1.
	\end{array}\right.
\end{equation}
Thus, the normalized posteriors are given by
\begin{equation}
	\label{eqn:posteriors}
	\begin{split}
x_t(0 | y) &= \frac{e_y [a r + (1-b) s]}{e_y [a r + (1-b) s] + f_y [(1-a) r + b s]}, \\
 x_t(1 | y) &= \frac{f_y [(1-a) r + b s]}{e_y [a r + (1-b) s] + f_y [(1-a) r + b s]}.
 	\end{split}
\end{equation}
\smallskip

\noindent\textbf{Step 2:} In this step, we 
compute $\hat{\rho}' = \text{Tr}_{\hat{\rho}_{t-1}} [\hat{U}_1 (\hat{\rho}_{t-1} \otimes \hat{\rho}_{\text{env}}^{(1)}) \hat{U}_1^\dagger].$
Firstly,
\begin{align*}
\hat{\rho}_{t-1} \otimes \hat{\rho}_{\text{env}}^{(1)} = \begin{pmatrix} r & c \\ c^* & s \end{pmatrix} \otimes \begin{pmatrix} 1 & 0 \\ 0 & 0 \end{pmatrix} = \begin{pmatrix} r & 0 & c & 0 \\ 0 & 0 & 0 & 0 \\ c^* & 0 & s & 0 \\ 0 & 0 & 0 & 0 \end{pmatrix},
\end{align*}
which can be equivalently written as
\begin{align}
	\label{eqn:first_product}
\hat{\rho}_{t-1} \otimes \hat{\rho}_{\text{env}}^{(1)} =&r |0_S\rangle\langle 0_S| \otimes |0_E\rangle\langle 0_E| + s |1_S\rangle\langle 1_S| \otimes |0_E\rangle\langle 0_E| \\
&+ c |0_S\rangle\langle 1_S| \otimes |0_E\rangle\langle 0_E| + c^* |1_S\rangle\langle 0_S| \otimes |0_E\rangle\langle 0_E|,\nonumber
\end{align}
with $S$ stands for the system particle and $E$ stands for the environment particle.

Design the transition unitary as
\begin{align*}
	\hat{U}_1 = \begin{pmatrix} \sqrt{a} & \sqrt{1-a} & 0 & 0 \\ \sqrt{1-a} & -\sqrt{a} & 0 & 0 \\ 0 & 0 & \sqrt{1-b} & \sqrt{b} \\ 0 & 0 & \sqrt{b} & -\sqrt{1-b} \end{pmatrix},
\end{align*}
which can be written as $\hat{U}_1 = U_1 \oplus U_2$ with
\begin{align*}
	U_1 = \begin{pmatrix} \sqrt{a} & \sqrt{1-a} \\ \sqrt{1-a} & -\sqrt{a} \end{pmatrix} \quad\text{and}\quad U_2 = \begin{pmatrix} \sqrt{1-b} & \sqrt{b} \\ \sqrt{b} & -\sqrt{1-b} \end{pmatrix},
\end{align*}
or simply
$$\hat{U}_1 = |0_S\rangle\langle 0_S| \otimes U_1 + |1_S\rangle\langle 1_S| \otimes U_2.$$

Applying $\hat{U}_1$ to both sides of $\hat{\rho}_{t-1} \otimes \hat{\rho}_{\text{env}}^{(1)}$ in equation \eqref{eqn:first_product} yields that
\begin{align}
\label{eqn:first_product_U1}
&\hat{U}_1 (\hat{\rho}_{t-1} \otimes \hat{\rho}_{\text{env}}^{(1)}) \hat{U}_1^\dagger\\
&= r |0_S\rangle\langle 0_S| \otimes (U_1 |0_E\rangle\langle 0_E| U_1^\dagger)+s |1_S\rangle\langle 1_S| \otimes (U_2 |0_E\rangle\langle 0_E| U_2^\dagger)\nonumber\\
&\quad+ c |0_S\rangle\langle 1_S| \otimes (U_1 |0_E\rangle\langle 0_E| U_1^\dagger) + c^* |1_S\rangle\langle 0_S| \otimes (U_2 |0_E\rangle\langle 0_E| U_2^\dagger),\nonumber
\end{align}
where we used $\langle 0_S |1_S\rangle=\langle 1_S |0_S\rangle=0$.  
Note that 
$$
U_1 |0_E\rangle = \sqrt{a} |0_E\rangle + \sqrt{1-a} |1_E\rangle =: |\phi_1\rangle,
$$
and hence
$$
U_1 |0_E\rangle\langle 0_E| U_1^\dagger = |\phi_1\rangle\langle \phi_1| = \begin{pmatrix} a & \sqrt{a(1-a)} \\ \sqrt{a(1-a)} & 1-a \end{pmatrix}.
$$
Note that 
$$
U_2 |0_E\rangle = \sqrt{1-b} |0_E\rangle + \sqrt{b} |1_E\rangle =: |\phi_2\rangle,
$$
and hence
$$
U_2 |0_E\rangle\langle 0_E| U_2^\dagger = |\phi_2\rangle\langle \phi_2| = \begin{pmatrix} 1-b & \sqrt{(1-b)b} \\ \sqrt{(1-b)b} & b \end{pmatrix}.
$$
Plugging the above results into equation \eqref{eqn:first_product_U1}, we have
\begin{align*}
	\hat{U}_1 (\hat{\rho}_{t-1} \otimes \hat{\rho}_{\text{env}}^{(1)}) \hat{U}_1^\dagger
&= r |0_S\rangle\langle 0_S| \otimes |\phi_1\rangle\langle \phi_1| + s |1_S\rangle\langle 1_S| \otimes |\phi_2\rangle\langle \phi_2| \\
&\quad+ c |0_S\rangle\langle 1_S| \otimes |\phi_1\rangle\langle \phi_2| + c^* |1_S\rangle\langle 0_S| \otimes |\phi_2\rangle\langle \phi_1|.
\end{align*}

Now, we work on the partial trace
\begin{align*}
\hat{\rho}' &= \text{Tr}_{\hat{\rho}_{t-1}} [\hat{U}_1 (\hat{\rho}_{t-1} \otimes \hat{\rho}_{\text{env}}^{(1)}) \hat{U}_1^\dagger]\\
&= \sum_{i=0,1} \langle i_S | \hat{U}_1 (\hat{\rho}_{t-1} \otimes \hat{\rho}_{\text{env}}^{(1)}) \hat{U}_1^\dagger | i_S \rangle\\
&=r\langle 0_S |0_S\rangle\langle 0_S| 0_S\rangle\otimes |\phi_1\rangle\langle \phi_1| +s\langle 1_S |1_S\rangle\langle 1_S| 1_S\rangle\otimes |\phi_2\rangle\langle \phi_2| \\
&= r |\phi_1\rangle\langle \phi_1| + s |\phi_2\rangle\langle \phi_2| \\
&= \begin{pmatrix} r a + s (1-b) & r \sqrt{a(1-a)} + s \sqrt{(1-b)b} \\ r \sqrt{a(1-a)} + s \sqrt{(1-b)b} & r (1-a) + s b \end{pmatrix},
\end{align*}
where we used $\langle 0_S |1_S\rangle=\langle 1_S |0_S\rangle=0$ in obtaining the third equation. Note that the diagonal of $\hat{\rho}'$ is $(r a + s (1-b), r (1-a) + s b)$, matching $A x_{t-1}$.
\medskip

\noindent\textbf{Step 3:} In this step, we compute $\hat{U}_2 \left( \hat{\rho}' \otimes \hat{\rho}_{\text{env}}^{(2)} \right) \hat{U}_2^\dagger$. 
Recall that $\hat{\rho}_{\text{env}}^{(2)}$ is defined as
$$
\hat{\rho}_{\text{env}}^{(2)} = |0\rangle\langle 0| = \begin{pmatrix} 1 & 0 & 0 \\ 0 & 0 & 0 \\ 0 & 0 & 0 \end{pmatrix}.
$$
Define the unitary $\hat{U}_2$ as $
\hat{U}_2 = V_0 \oplus V_1$, where 
$$V_0 = \begin{pmatrix} \sqrt{e_0} & -\frac{\sqrt{e_1}}{\sqrt{e_0} \sqrt{1 + \frac{e_1}{e_0}}} & \frac{ -\frac{\sqrt{e_2}}{\sqrt{e_0}} + \frac{e_1 \sqrt{e_2}}{e_0^{3/2} (1 + \frac{e_1}{e_0})} }{ \sqrt{ \left( \frac{\sqrt{e_2}}{\sqrt{e_0}} - \frac{e_1 \sqrt{e_2}}{e_0^{3/2} (1 + \frac{e_1}{e_0})} \right)^2 + 1 + \frac{e_1 e_2}{e_0^2 (1 + \frac{e_1}{e_0})^2} } } \vspace{0.25cm}\\ \sqrt{e_1} & \frac{1}{\sqrt{1 + \frac{e_1}{e_0}}} & -\frac{\sqrt{e_1} \sqrt{e_2}}{e_0 (1 + \frac{e_1}{e_0}) \sqrt{ \left( \frac{\sqrt{e_2}}{\sqrt{e_0}} - \frac{e_1 \sqrt{e_2}}{e_0^{3/2} (1 + \frac{e_1}{e_0})} \right)^2 + 1 + \frac{e_1 e_2}{e_0^2 (1 + \frac{e_1}{e_0})^2} } } \vspace{0.25cm}\\ \sqrt{e_2} & 0 & \frac{1}{\sqrt{ \left( \frac{\sqrt{e_2}}{\sqrt{e_0}} - \frac{e_1 \sqrt{e_2}}{e_0^{3/2} (1 + \frac{e_1}{e_0})} \right)^2 + 1 + \frac{e_1 e_2}{e_0^2 (1 + \frac{e_1}{e_0})^2} }} \end{pmatrix}, $$

$$V_1 =  \begin{pmatrix} \sqrt{f_0} & -\frac{\sqrt{f_1}}{\sqrt{f_0} \sqrt{1 + \frac{f_1}{f_0}}} & \frac{ -\frac{\sqrt{f_2}}{\sqrt{f_0}} + \frac{f_1 \sqrt{f_2}}{f_0^{3/2} (1 + \frac{f_1}{f_0})} }{ \sqrt{ \left( \frac{\sqrt{f_2}}{\sqrt{f_0}} - \frac{f_1 \sqrt{f_2}}{f_0^{3/2} (1 + \frac{f_1}{f_0})} \right)^2 + 1 + \frac{f_1 f_2}{f_0^2 (1 + \frac{f_1}{f_0})^2} } } \vspace{0.25cm}\\ \sqrt{f_1} & \frac{1}{\sqrt{1 + \frac{f_1}{f_0}}} & -\frac{\sqrt{f_1} \sqrt{f_2}}{f_0 (1 + \frac{f_1}{f_0}) \sqrt{ \left( \frac{\sqrt{f_2}}{\sqrt{f_0}} - \frac{f_1 \sqrt{f_2}}{f_0^{3/2} (1 + \frac{f_1}{f_0})} \right)^2 + 1 + \frac{f_1 f_2}{f_0^2 (1 + \frac{f_1}{f_0})^2} } } \vspace{0.25cm}\\ \sqrt{f_2} & 0 & \frac{1}{\sqrt{ \left( \frac{\sqrt{f_2}}{\sqrt{f_0}} - \frac{f_1 \sqrt{f_2}}{f_0^{3/2} (1 + \frac{f_1}{f_0})} \right)^2 + 1 + \frac{f_1 f_2}{f_0^2 (1 + \frac{f_1}{f_0})^2} }} \end{pmatrix}.
$$
We notice that
$$
V_0 |0\rangle = \sqrt{e_0} |0\rangle + \sqrt{e_1} |1\rangle + \sqrt{e_2} |2\rangle\quad \text{and}\quad V_1 |0\rangle = \sqrt{f_0} |0\rangle + \sqrt{f_1} |1\rangle + \sqrt{f_2} |2\rangle.
$$
The $2\times 2$-dimensional density matrix $\hat{\rho}'$ can be expressed as
$$\hat{\rho}' = \sum_{j,k=0}^1 \rho'_{jk} |j\rangle\langle k|, \qquad \rho'_{jk} = \langle j | \hat{\rho}' | k \rangle.$$
Consequently,
$$
\hat{U}_2 (\hat{\rho}' \otimes |0\rangle\langle 0|) \hat{U}_2^\dagger = \sum_{j,k=0}^1 \rho'_{jk} |j\rangle\langle k| \otimes (V_j |0\rangle)(V_k |0\rangle)^\dagger.
$$

\noindent\textbf{Step 4:} In this step, we compute the full expression $$\text{Tr}_{\hat{\rho}_{\text{env}}^{(2)}} \left( \hat{P}_y \hat{U}_2 \left( \hat{\rho}' \otimes \hat{\rho}_{\text{env}}^{(2)} \right) \hat{U}_2^\dagger \hat{P}_y \right).$$
Applying the projector $\hat{P}_y = I_2 \otimes |y\rangle\langle y|$ gives
$$\hat{P}_y \hat{U}_2 (\hat{\rho}' \otimes |0\rangle\langle 0|) \hat{U}_2^\dagger \hat{P}_y = \sum_{j,k=0}^1 \rho'_{jk} |j\rangle\langle k| \otimes |y\rangle \langle y | V_j |0\rangle \langle 0 | V_k^\dagger | y \rangle \langle y |.$$
Since $\langle 0 | V_k^\dagger | y \rangle = \langle y | V_k |0\rangle^*$, we have
$$\hat{P}_y \hat{U}_2 (\hat{\rho}' \otimes |0\rangle\langle 0|) \hat{U}_2^\dagger \hat{P}_y =\sum_{j,k=0}^1 \rho'_{jk} \langle y | V_j |0\rangle \langle y | V_k |0\rangle^* |j\rangle\langle k| \otimes |y\rangle\langle y|.$$
Since
$$\langle y | V_1 |0\rangle = \sqrt{e_y} \quad \text{and}\quad \langle y | V_2 |0\rangle = \sqrt{f_y},$$
taking partial trace yields
$$\text{Tr}_{\hat{\rho}_{\text{env}}^{(2)}} \left( \hat{P}_y \hat{U}_2 \left( \hat{\rho}' \otimes \hat{\rho}_{\text{env}}^{(2)} \right) \hat{U}_2^\dagger \hat{P}_y \right)  =  \sum_{j,k=0}^1 \rho'_{jk} \langle y | V_j |0\rangle \langle y | V_k^\dagger |0\rangle |j\rangle\langle k|.$$
By \eqref{eqn:HMM_quantum}, we have
$$\text{diag}(\hat{\rho}_t) \propto \begin{pmatrix} (r a + s (1-b)) e_y \vspace{0.1cm}\\ (r (1-a) + s b) f_y \end{pmatrix}.$$
The normalization constant of $\hat{\rho}_t$ is given by
$$\text{Tr}(\hat{\rho}_t) = e_y [a r + (1-b) s] + f_y [(1-a) r + b s].$$
Therefore, the diagonal of the normalized $\hat{\rho}_t$
$$\left(\frac{e_y [a r + (1-b) s]}{\text{Tr}(\hat{\rho}_t)}, \quad \frac{f_y [(1-a) r + b s]}{\text{Tr}(\hat{\rho}_t)}\right)$$
coincides with the posterior distribution of the HMM obtained in equation \eqref{eqn:posteriors}.
\end{proof}

\bibliography{bib-ms}

\begin{thebibliography}{}

\bibitem[Adhikary et~al., 2020]{adhikary2020expressiveness}
Adhikary, S., Srinivasan, S., Gordon, G., and Boots, B. (2020).
\newblock Expressiveness and learning of hidden quantum {M}arkov models.
\newblock In {\em International Conference on Artificial Intelligence and
  Statistics}, pages 4151--4161. PMLR.

\bibitem[Aliakbarpour et~al., 2025]{aliakbarpour2025adversarially}
Aliakbarpour, M., Braverman, V., Chia, N.-H., and Liu, Y. (2025).
\newblock Adversarially robust quantum state learning and testing.
\newblock {\em arXiv preprint arXiv:2508.13959}.

\bibitem[Cholewa et~al., 2017]{cholewa2017quantum}
Cholewa, M., Gawron, P., G{\l}omb, P., and Kurzyk, D. (2017).
\newblock Quantum hidden {M}arkov models based on transition operation
  matrices.
\newblock {\em Quantum Information Processing}, 16(4):101.

\bibitem[Clark et~al., 2015]{clark2015open}
Clark, J., Zhang, J., and Wiesner, K. (2015).
\newblock Hidden quantum {M}arkov models and open quantum systems with
  instantaneous feedback.
\newblock {\em Proceedings of the Royal Society A: Mathematical, Physical and
  Engineering Sciences}, 471(2178):20150320.

\bibitem[Fran\c{c}a et~al., 2021]{brandao2020fast}
Fran\c{c}a, D.~S., Brand\~{a}o, F. G.~L., and Kueng, R. (2021).
\newblock {Fast and Robust Quantum State Tomography from Few Basis
  Measurements}.
\newblock In Hsieh, M.-H., editor, {\em 16th Conference on the Theory of
  Quantum Computation, Communication and Cryptography (TQC 2021)}, volume 197
  of {\em Leibniz International Proceedings in Informatics (LIPIcs)}, pages
  7:1--7:13, Dagstuhl, Germany. Schloss Dagstuhl -- Leibniz-Zentrum f{\"u}r
  Informatik.

\bibitem[Jaeger, 2000]{jaeger2000observable}
Jaeger, H. (2000).
\newblock Observable operator models for discrete stochastic time series.
\newblock {\em Neural computation}, 12(6):1371--1398.

\bibitem[Javidian et~al., 2021]{javidian2021chqmm}
Javidian, M., Aggarwal, V., and Jacob, A.~K. (2021).
\newblock Learning circular hidden quantum {M}arkov models: A tensor network
  approach.
\newblock {\em IEEE Transactions on Quantum Engineering}, 2:1--12.
\newblock Also available at \url{https://arxiv.org/abs/2111.01536}.

\bibitem[Li et~al., 2023]{li2023shqmm}
Li, X., Zhu, Y., Wu, Y., Yang, X., Yu, J., and Chen, X. (2023).
\newblock Split hidden quantum {M}arkov model inspired by quantum conditional
  master equation.
\newblock {\em Quantum Information Processing}, 22(3):107.

\bibitem[Ma et~al., 2023]{ma2023financial}
Ma, H., Zhou, S., Li, L., and Zhang, J. (2023).
\newblock Hidden quantum {M}arkov models in financial time series: A
  quantum-inspired perspective.
\newblock {\em Mathematics}, 13(15):2505.

\bibitem[Monras et~al., 2010]{monras2010hidden}
Monras, A., Beige, A., and Wiesner, K. (2010).
\newblock Hidden quantum {M}arkov models and non-adaptive read-out of many-body
  states.
\newblock {\em arXiv preprint arXiv:1002.2337}.

\bibitem[Ning and Ionides, 2023]{ning2023iterated}
Ning, N. and Ionides, E.~L. (2023).
\newblock Iterated block particle filter for high-dimensional parameter
  learning: Beating the curse of dimensionality.
\newblock {\em Journal of Machine Learning Research}, 24(82):1--76.

\bibitem[O’Neill et~al., 2012]{oneill2012toy}
O’Neill, B., Clark, J., and Wiesner, K. (2012).
\newblock Simple examples of hidden quantum {M}arkov models.
\newblock {\em Journal of Physics A: Mathematical and Theoretical},
  45(9):095305.

\bibitem[Preskill, 2018]{preskill2018quantum}
Preskill, J. (2018).
\newblock Quantum computing in the {NISQ} era and beyond.
\newblock {\em Quantum}, 2:79.

\bibitem[Qiu et~al., 2018]{qiu2018multivariate}
Qiu, J., Jammalamadaka, S.~R., and Ning, N. (2018).
\newblock Multivariate {B}ayesian structural time series model.
\newblock {\em Journal of Machine Learning Research}, 19(68):1--33.

\bibitem[Srinivasan et~al., 2018a]{srinivasan2018hilbert}
Srinivasan, S., Downey, C., and Boots, B. (2018a).
\newblock Learning and inference in hilbert space with quantum graphical
  models.
\newblock In {\em Advances in Neural Information Processing Systems},
  volume~31. Curran Associates, Inc.

\bibitem[Srinivasan et~al., 2018b]{srinivasan2018learning}
Srinivasan, S., Gordon, G., and Boots, B. (2018b).
\newblock Learning hidden quantum {M}arkov models.
\newblock In {\em International Conference on Artificial Intelligence and
  Statistics}, pages 1979--1987. PMLR.

\bibitem[Stilck~Fran{\c{c}}a et~al., 2024]{stilck2024efficient}
Stilck~Fran{\c{c}}a, D., {M}arkovich, L.~A., Dobrovitski, V., Werner, A.~H.,
  and Borregaard, J. (2024).
\newblock Efficient and robust estimation of many-qubit {H}amiltonians.
\newblock {\em Nature Communications}, 15(1):311.

\bibitem[Yu et~al., 2023]{yu2023robust}
Yu, W., Sun, J., Han, Z., and Yuan, X. (2023).
\newblock Robust and efficient {H}amiltonian learning.
\newblock {\em Quantum}, 7:1045.

\end{thebibliography}

\end{document}